\documentclass{article}

\usepackage{arxiv}
\usepackage[utf8]{inputenc} 
\usepackage[T1]{fontenc}    
\usepackage{hyperref}       
\usepackage{url}            
\usepackage{booktabs}       
\usepackage{amsfonts}       
\usepackage{nicefrac}       
\usepackage{microtype}      
\usepackage{lipsum}		    
\usepackage{graphicx}
\usepackage[square,numbers]{natbib} 
\usepackage{doi}
\usepackage{kotex} 
\usepackage{amsmath} 
\usepackage{multirow}
\usepackage[left]{lineno} 
\DeclareMathOperator*{\argmin}{argmin} 

\begin{document}


\title{AI-powered Digital Twin of the Ocean: Reliable Uncertainty Quantification for Real-time Wave Height Prediction with Deep Ensemble}

\date{}                     


\author{
  Dongeon Lee$^{a}$, Sunwoong Yang$^{a}$, Jae-Won Oh$^{b}$, Su-Gil Cho$^{b}$, Sanghyuk Kim$^{c}$, Namwoo Kang$^{a,d,}$\thanks{Corresponding author: nwkang@kaist.ac.kr} \\\\
  \textit{$^{a}$Cho Chun Shik Graduate School of Mobility, Korea Advanced Institute of Science and Technology,} \\
  \textit{Daejeon, 34051, Republic of Korea} \\
  \textit{$^{b}$Eco-friendly Ocean Development Research Division, Korea Research Institute of Ships and Ocean Engineering,} \\
  \textit{Daejeon, 34103, Republic of Korea} \\
  \textit{$^{c}$Department of Mechanical Engineering, Korea Advanced Institute of Science and Technology,} \\
  \textit{Daejeon, 34141, Republic of Korea} \\
  \textit{$^{d}$Narnia Labs, Daejeon, 34051, Republic of Korea}
}

\renewcommand{\headeright}{}
\renewcommand{\undertitle}{}
\renewcommand{\shorttitle}{}

\setlength{\parindent}{15pt} 

\hypersetup{
pdftitle={A template for the arxiv style},
pdfsubject={q-bio.NC, q-bio.QM},
pdfauthor={David S.~Hippocampus, Elias D.~Striatum},
pdfkeywords={First keyword, Second keyword, More},
}

\maketitle

\begin{abstract}
Environmental pollution and fossil fuel depletion have prompted the need for renewable energy-based power generation. However, its stability is often challenged by low energy density and non-stationary conditions. Wave energy converters (WECs), in particular, need reliable real-time wave height prediction to address these issues caused by irregular wave patterns, which can lead to the inefficient and unstable operation of WECs. In this study, we propose an AI-powered reliable real-time wave height prediction model that integrates long short-term memory (LSTM) networks for temporal prediction with deep ensemble (DE) for robust uncertainty quantification (UQ), ensuring high accuracy and reliability. To further enhance the reliability, uncertainty calibration is applied, which has proven to significantly improve the quality of the quantified uncertainty. Using real operational data from an oscillating water column-wave energy converter (OWC-WEC) system in Jeju, South Korea, the model achieves notable accuracy ($R^2 > 0.9$), while increasing uncertainty quality by over 50\% through simple calibration technique. Furthermore, a comprehensive parametric study is conducted to explore the effects of key model hyperparameters, offering valuable guidelines for diverse operational scenarios, characterized by differences in wavelength, amplitude, and period. These results demonstrate the model's capability to deliver reliable predictions, facilitating digital twin of the ocean.
\end{abstract}

\keywords{Digital twin \and Wave energy converter \and Wave height prediction \and Uncertainty quantification \and Uncertainty calibration \and Deep ensemble}



\section{Introduction}
\label{sec:introduction}
Energy decarbonization and sustainable industries are essential for addressing global climate change, as emphasized in the sustainable development goals (SDGs) adopted by all United Nations Member States in 2015 \cite{colglazier2015sustainable, sachs2019six}. These goals highlight the transformation from conventional fossil fuels to renewable energy resources, which are capable of reducing greenhouse gas emissions and mitigating climate change. From this perspective, the utilization of renewable energy resources and technologies is continuously required. The most widely used renewable energy types are wind energy, solar energy, and hydropower. Especially, ocean waves generated by wind, with their large potential energy capacity and higher energy density compared to other renewable resources \cite{rehman2023review}, are expected to play a significant role in achieving SDG 7 (Affordable and Clean Energy) and SDG 13 (Climate Action). Consequently, global interest in wave energy-based power generation has been steadily increasing over time \cite{khojasteh2023large, li2024advanced}.

In perspective of wave energy conversion technology, various systems called WECs have been proposed and have experienced rapid growth in recent decades to utilize the enormous energy resources in the ocean \citep{falcao2010wave, khaligh2017energy, aderinto2018ocean, wendt2019ocean}. Primary wave energy conversion is achieved by oscillating water column-based systems, including floating body, oscillating solid member, and oscillating water within a structure \cite{brooke2003wave, mccormick2013ocean, babarit2017ocean}. Unlike hydropower energy, the power generation using WECs mainly depends on short-term and local conditions such as wave height, period, and spectra. It means that these systems may face several limitations in stable power generation due to local and irregular wave patterns. Therefore, real-time wave height prediction is essential to increase the availability on WECs. In fact, recent studies have demonstrated the importance of real-time wave height prediction model for the operation of the WECs \cite{fusco2010short, fusco2011study, li2012wave, son2017optimizing, previsic2021ocean, abad2024experimental}. Moreover, these prediction models are also necessary to construct digital twins of the WECs, which are virtual representations of physical objects and systems, to make real-time decisions regarding the stable operation.

There have been numerous studies on wave height prediction, focusing on the development of theoretical and statistical models \cite{kinsman1984wind, komen1984existence, stoker1992water,komen1996dynamics, altunkaynak2005significant, reikard2011forecasting, falnes2020ocean}. More recently, machine learning (ML)-based approaches, including deep learning (DL), have been increasingly used to overcome the limitations of conventional methodologies, particularly in handling nonlinear patterns and large datasets \cite{malekmohamadi2011evaluating, kollisch2018nonlinear, james2018machine, fan2020novel, huang2021improved, bento2021ocean, mahdavi2023application, zilong2022spatial, xu2023instantaneous, le2023prediction, cheng2023prediction, pang2023novel, liu2024phase, mahdavi2024development, lei2024wave}. These studies involve various datasets provided by simulation, experiment, operational, and public data repositories, covering scales from local to global. An oscillating water column-wave energy converter (OWC-WEC), which is the primary focus of this study, also requires high-quality and reliable wave height prediction models for its operation and management. Various studies have explored numerical and experimental approaches \cite{kim2020numerical, kim2021numerical, orphin2021uncertainty, kim2023experimental, ning2023experimental}, as well as artificial intelligence (AI)-based approach using the LSTM algorithm to handle time-series wave patterns data \cite{seo2021prediction}. However, most existing wave height prediction models are limited in that they cannot provide the information on the uncertainty over their predictions. More specifically, their deterministic models provide a single prediction value, while probabilistic models provide not only the prediction values, but also the quantified uncertainty in their predictions. In real-world applications, deterministic prediction models cannot provide uncertainty estimates, which may lead to catastrophic outcomes in WEC operations, such as excessive air inflow resulting from unanticipated changes in wave height that the deterministic model fails to account for, potentially damaging the air turbine.

From a DL perspective on UQ, there are two main categories \cite{abdar2021review}: Bayesian-based approach \cite{gal2016dropout, kendall2017uncertainties} and ensemble-based approach \cite{lakshminarayanan2017simple}. The Bayesian-based approach, commonly known as Bayesian neural network (BNN) has been widely used for UQ in DL models \cite{jospin2022hands}. For real-time ocean wave prediction, a BNN-based prediction method was proposed in \cite{zhang2022phase}. On the other hand, the ensemble-based approach, known as DE, is also often used to effectively calculate uncertainty owing to their advantages: simplicity and scalability. In particular, DE is considered more practical than BNN for dataset shifting, as it explores diverse modes in function space, whereas Bayesian-based approach tends to fall into a single mode \cite{fort2019deep}. Despite these advantages, there are no known applications of DE in wave height prediction, especially for short-term and local scales, although research has been conducted on probabilistic spatiotemporal forecasting for solar irradiation on a global scale \cite{liu2021probabilistic}. Furthermore, in uncertainty estimation for AI-based classification and regression models, including DL, it is crucial to conduct quantitative quality evaluations to ensure the reliability of the probabilistic model and apply uncertainty calibrations based on these evaluations metrics \cite{naeini2015obtaining, guo2017calibration, kuleshov2018accurate, gustafsson2020evaluating, scalia2020evaluating, levi2022evaluating}. However, prior studies, including the BNN-based method for wave height prediction \cite{zhang2022phase}, have not incorporated calibration methods to enhance the quality of UQ. In this regard, the calibration framework for DEs have been proposed in multi-output regression tasks \cite{yang2024towards} to obtain reliable probabilistic predictions. Although this research represents the state-of-the-art reliable probabilistic predictions in multi-output regression tasks, which could be applicable to time-series data like wave patterns, it has not yet been validated for temporal multi-output regression tasks.

\begin{figure*}[htb!]
    \centering
        \includegraphics[width=.95\textwidth]{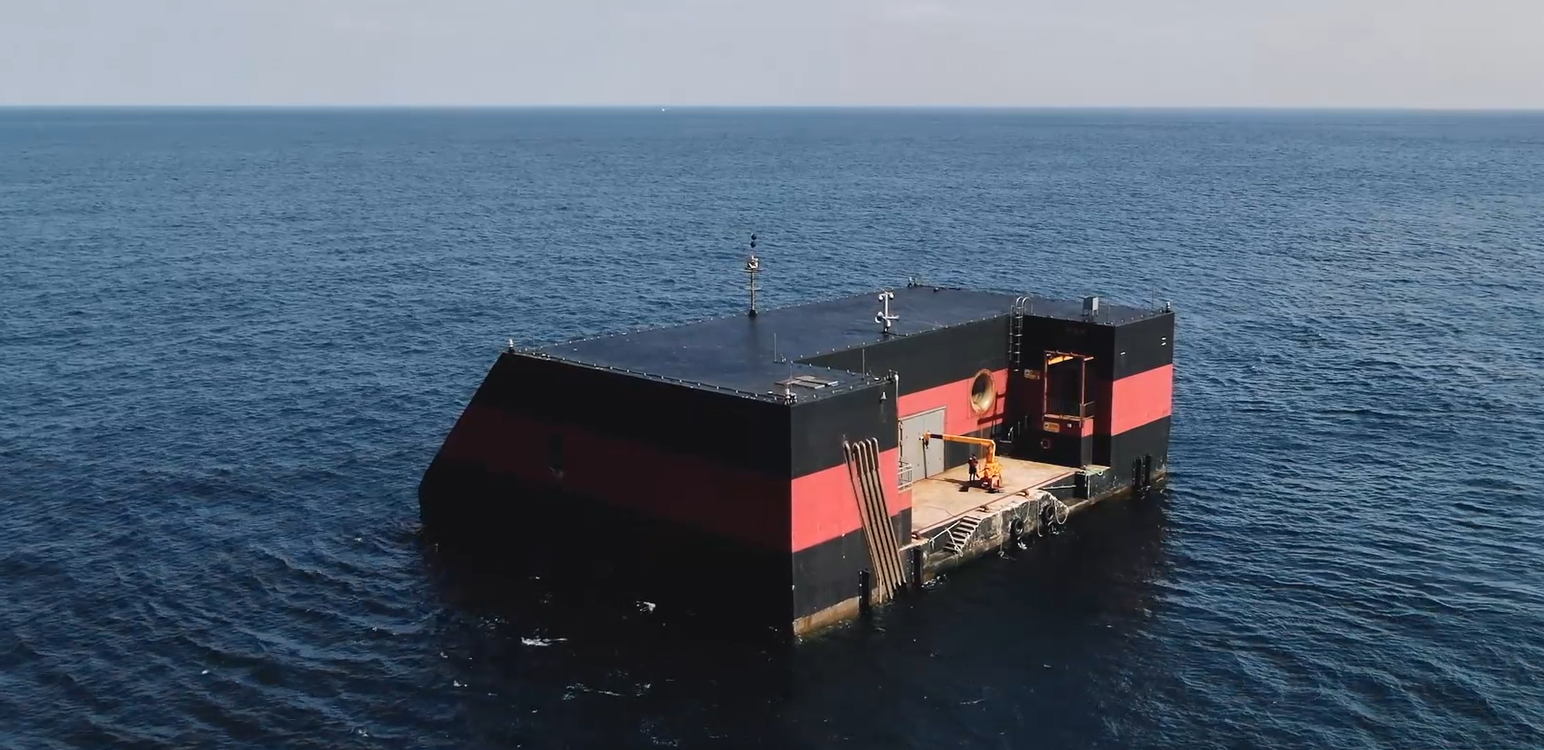}      
    \caption{500kW OWC-WEC at Yongsoo, Jeju, South Korea.}
    \label{fig:owc_wec_jeju}
\end{figure*}

\begin{figure*}[htb!]
    \centering
        \includegraphics[width=.95\textwidth]{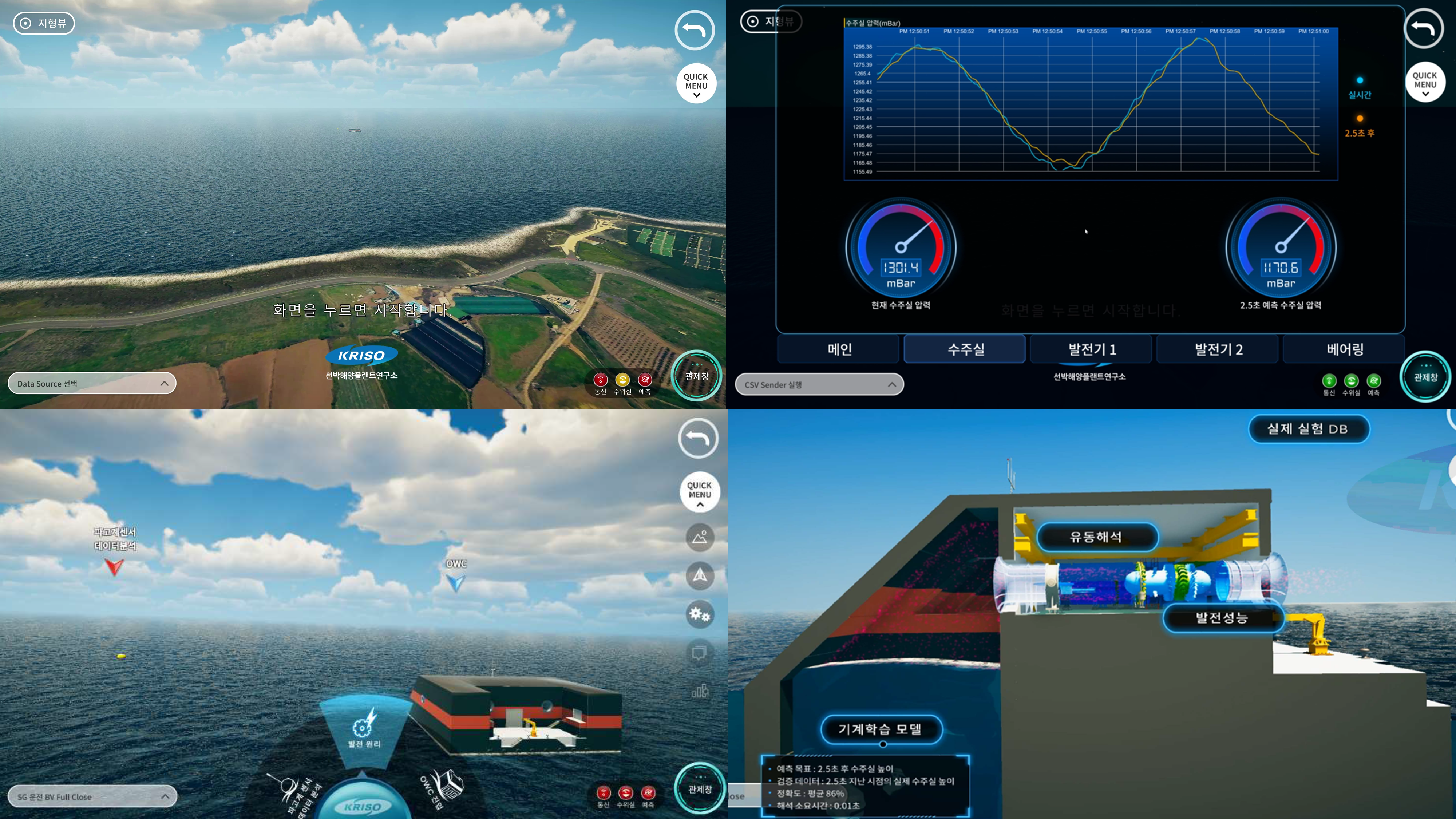}        
    \caption{Digital twin of OWC-WEC developed by KRISO.}
    \label{fig:dt_kriso}
\end{figure*}

In this study, we propose a simple and scalable LSTM-DE model architecture to construct a reliable real-time wave height prediction model to be embedded into digital twin of WECs. Specifically, we conduct the research mainly considering the following three aspects: (1) a simplicity and scalability of LSTM-DE model architecture, (2) high accuracy for prediction, and (3) reliable UQ coupled with uncertainty calibration. Within these perspectives, we also demonstrate the impact of ocean-related domain knowledge on model performance to implement effective and efficient experimental design and optimize the hyperparameters of DL model architectures. To verify the effectiveness and applicability of embedding into digital twin of WECs, we use real operational data from the 500kw OWC-WEC installed on the coast of Jeju Island, South Korea (Fig. \ref{fig:owc_wec_jeju}) \cite{falcao2016oscillating}, along with its digital twin platform developed by Korea Research Institute of Ships \& Ocean Engineering (KRISO) (Fig. \ref{fig:dt_kriso}). To the best of our knowledge, this is the first case of using a calibrated DE methodology for real-time wave height prediction. The corresponding results are expected to enhance the availability of WECs and facilitate the development of a digital twin of the ocean, providing robust and reliable real-time wave height predictions in perspective of temporal multi-output regression tasks.

The main contributions and novelties of this paper are summarized as follows:
\begin{enumerate}
    \item This is the first study to apply DE methodology with uncertainty calibration for reliable real-time wave height prediction, ensuring accurate predictions and reliable quantified uncertainties, which can be embedded into the digital twin of the ocean to support real-time decision-making for stable WEC operations.
    \item A lightweight yet powerful LSTM architecture, achieving efficiency with a single layer for reduced computational cost and leveraging the entire hidden state information to enhance predictive accuracy. This approach effectively captures highly nonlinear wave behaviors while ensuring both performance and scalability.
    \item A simple and scalable DE methodology is combined with an effective post-hoc uncertainty calibration technique to estimate model uncertainty and enhance its reliability, and its effectiveness is verified qualitatively and quantitatively.
    \item The baseline LSTM-DE model, combining the temporal modeling strength of LSTM with the robust UQ of DE, is developed using real operational data from an ocean-installed OWC-WEC system, which exhibits greater irregularity than simulation and experimental datasets used in previous research. The study also investigates the impact of ocean-related domain knowledge in designing initial experiments to minimize training costs.
    \item A comprehensive parametric study is conducted on various model hyperparameters and ocean wave types, characterized by differences in wavelength, amplitude, and period. This study evaluates predictive accuracy and uncertainty quality while incorporating uncertainty calibration to enhance model reliability, providing valuable insights and guidance for future researchers.
\end{enumerate}

The remainder of this paper is organized as follows. Section \ref{sec:methodology} provides a brief overview of the OWC-WEC, LSTM, DE, uncertainty evaluation, and its calibration. Section \ref{sec:lstm_de_architecture} explains the proposed LSTM-DE model architecture, corresponding to temporal multi-output regression with uncertainty, for real-time wave height prediction. Section \ref{sec:results_and_discussion} presents the experimental settings, results, and discussion. Finally, Section \ref{sec:conclusions_and_future_work} offers our conclusions and suggests directions for future work.

\section{Methodology}
\label{sec:methodology}

\subsection{Oscillating water column-wave energy converter (OWC-WEC)}
\label{subsec:owc_wec}
\begin{figure*}[htb!]
    \centering
        \includegraphics[width=.95\textwidth]{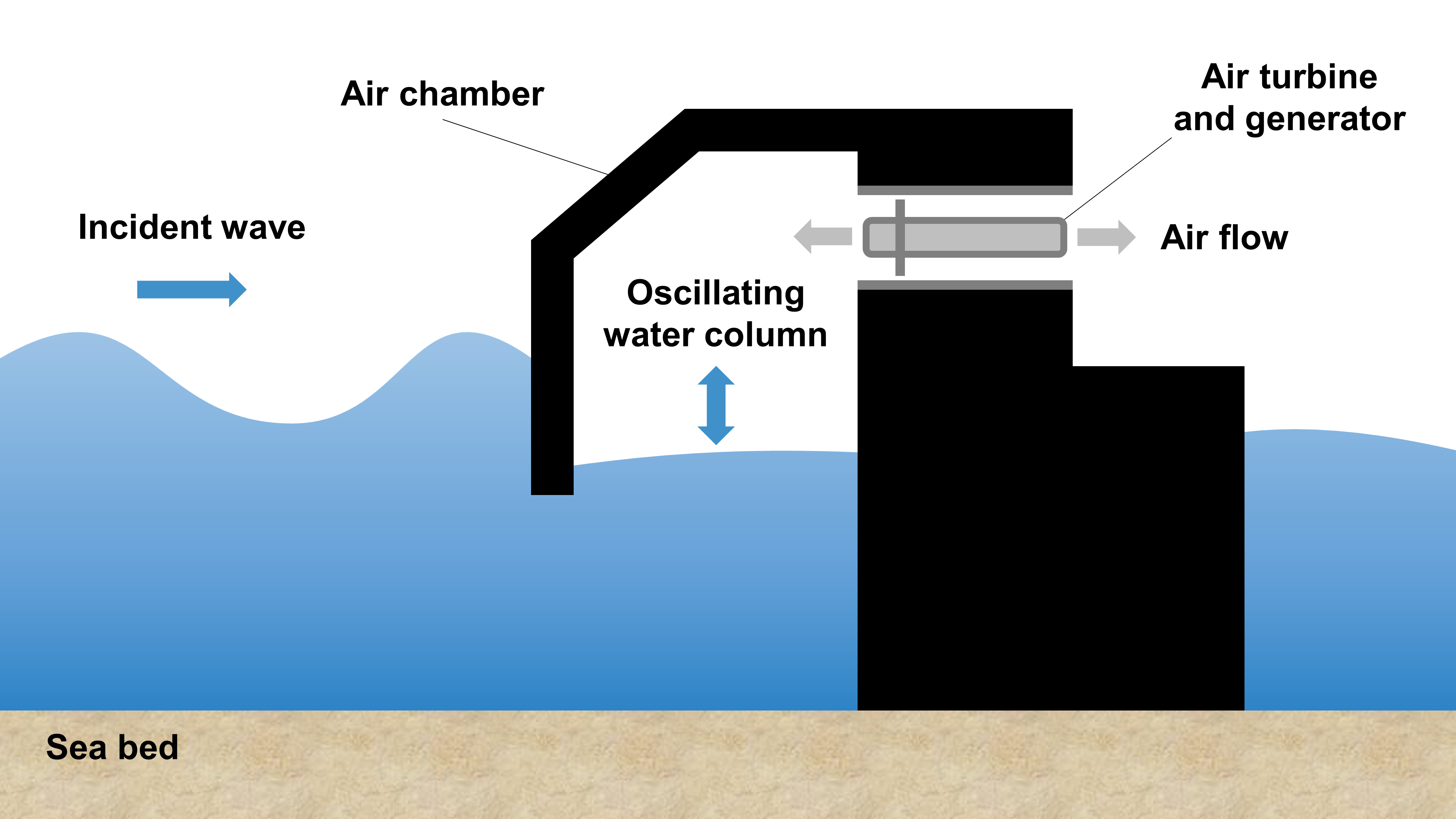}        
    \caption{Schematic diagram of OWC-WEC.}
    \label{fig:sd_owc_wec}
\end{figure*}

An oscillating water column-wave energy converter (OWC-WEC) is a renewable energy system designed to harness the vertical motion of ocean waves to generate electricity. The system consists of a partially submerged concrete structure, air chamber, air turbine, and generator, as shown in Fig. \ref{fig:sd_owc_wec}. When an incident wave interacts with the OWC-WEC, the oscillating motion of the water column within the air chamber forces the inside air. This movement generates an airflow that drives the connected air turbine. The rotation of turbine then produces electricity via the generator. With this regard, accurate and reliable wave predictions are essential for efficient operation and management of OWC-WEC, as wave behavior directly influences energy output and helps prevent system shutdowns, avoiding potential damage to the system. Moreover, the inherent variability of wave energy presents challenges to both operational efficiency and power generation stability. To address these challenges, it is crucial to predict the water column height inside the chamber in real time with both accurate predictions and reliable uncertainty estimates for efficient system control.

In this study, an LSTM-DE model architecture is proposed for real-time wave height prediction, using univariate time-series data collected from a pressure sensor installed at the bottom of the OWC-WEC chamber. LSTM is employed to effectively predict time-series data, while DE ensures simple and scalable UQ. Additionally, uncertainty calibration is applied to enhance model reliability further. The pressure data, convertible from pressure (mbar) to wave height (m) using the principles of hydrostatic pressure, enables the model to accurately capture the oscillating motion of the water column and reliably predict its behavior. By doing so, the system efficiently converts the ocean's kinetic energy into electrical energy, making it a valuable resource for ocean energy technologies.

\subsection{Long short-term memory (LSTM) network for temporal prediction}
\label{subsec:lstm}
In this study, a long short-term memory (LSTM) network was used to handle time-series data, such as temporal dynamics of ocean waves. The LSTM network is a type of recurrent neural network (RNN) \cite{pollack1990recursive, frasconi1998general}, designed to address the vanishing gradient problem, which affects long-term dependencies in traditional RNNs. By embedding a gating mechanism that allows information to be retained over extended time intervals, LSTM networks have been shown to learn long-term dependencies more effectively than simple RNN architectures \cite{bengio1994learning, hochreiter1997long, gers2000learning}, demonstrating successful capabilities in modeling sequential data and capturing long-term dependencies in various applications \cite{bengio2017deep, nielsen2019practical}.

\begin{figure*}[htb!]
    \centering
        \includegraphics[width=.95\textwidth]{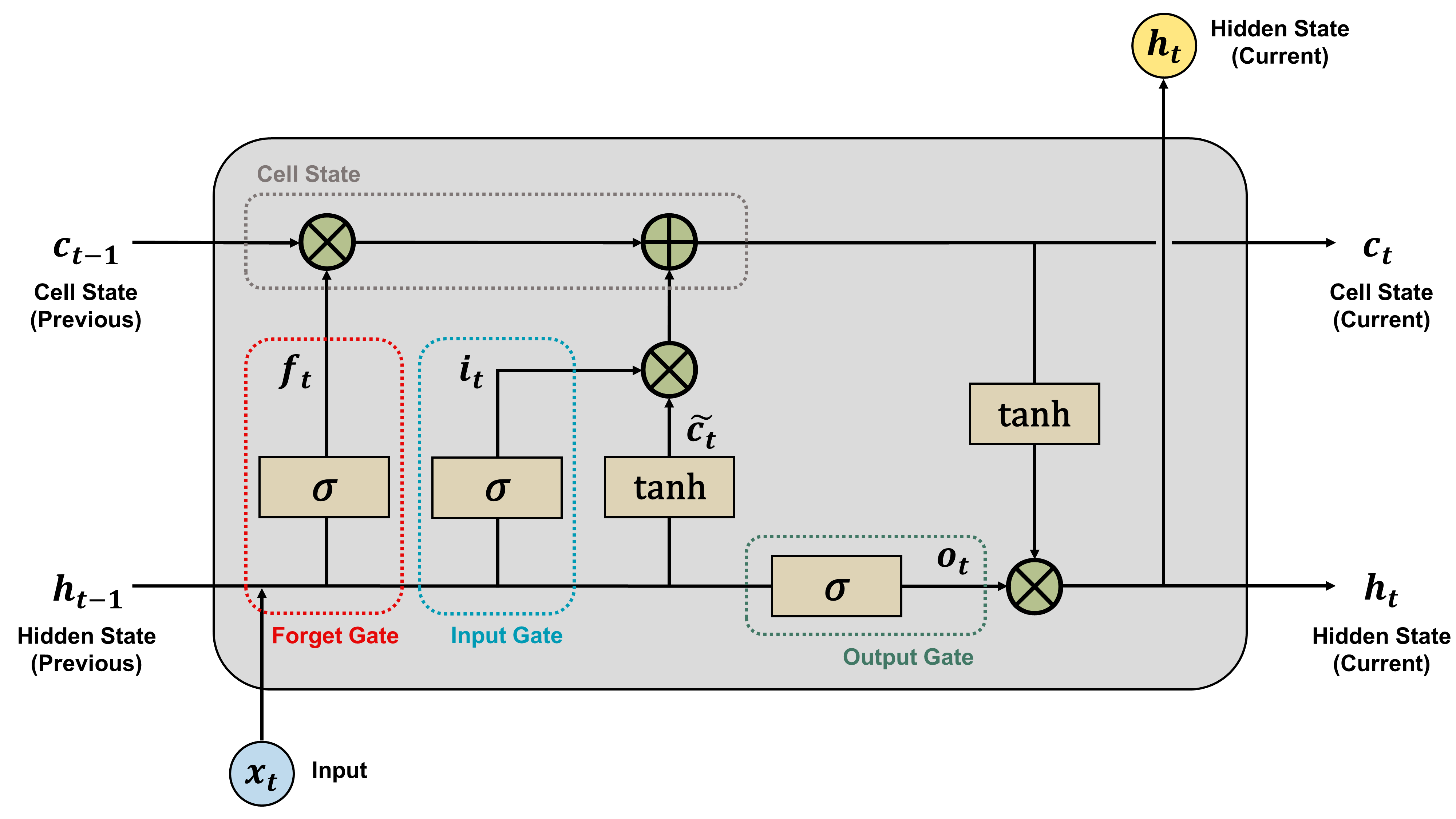}        
    \caption{Architecture of LSTM cell.}
    \label{fig:lstm_cell_diagram}
\end{figure*}

The main component of LSTM network is the LSTM cell, which consists of forget gate, input gate, and output gate, as shown in Fig. \ref{fig:lstm_cell_diagram}. These gates regulate the information of sequence data into, out of, and through the cell state, which stores the values as memories over various time intervals. This allows the LSTM network to retain information over long sequences, mitigating the vanishing gradient problem that traditional RNNs face. The key equations governing the LSTM cell are as follows:

\begin{equation}
\begin{aligned}
    f_t &= \sigma(w_f [h_{t-1}, x_t] + b_f) \\
    i_t &= \sigma(w_i [h_{t-1}, x_t] + b_i) \\
    \tilde{c}_t &= \tanh(w_c [h_{t-1}, x_t] + b_c) \\
    c_t &= f_t \odot c_{t-1} + i_t \odot \tilde{c}_t \\
    o_t &= \sigma(w_o [h_{t-1}, x_t] + b_o) \\
    h_t &= o_t \odot \tanh(c_t)
\end{aligned}
\label{eq:lstm_equations}
\end{equation}
where $x_t$ is the input at time $t$. $h_{t-1}$ and $h_t$ refer to the previous and current hidden states. $f_t$, $i_t$, and $o_t$ represent the forget, input, and output gates. $w$ and $b$ are the corresponding weights and biases for each gate. $c_t$ is the cell state, and $\tilde{c}_t$ is the candidate value. The function $\sigma(\cdot)$ is the sigmoid activation function, while $tanh(\cdot)$ is the hyperbolic tangent activation function. Finally, $\odot$ denotes element-wise multiplication.

The forget gate $f_t$ determines which information from the previous cell state $c_{t-1}$ should be retained or discarded, producing a value between 0 (forget) and 1 (preserve) through the sigmoid function $\sigma$, based on the previous hidden state $h_{t-1}$ and the current input $x_t$. Next, the input gate $i_t$ decides which new information will be added to the cell state $c_t$, while the $tanh$ layer generates a vector of candidate values $\tilde{c}_t$ that could be added to the cell state $c_t$. The cell state $c_t$ is updated by combining the previous cell state $c_{t-1}$, modulated by the forget gate $f_t$, with the candidate values $\tilde{c}_t$ from the input gate $i_t$, as shown in Eq. \eqref{eq:lstm_equations}. Finally, the output gate $o_t$ determines which part of the cell state will be output as the next hidden state $h_t$. The updated cell state $c_t$ is passed through the $tanh$ layer, and then multiplied by the output of the output gate $o_t$ to produce the hidden state $h_t$. This mechanism enables LSTM networks to effectively overcome the long-term dependencies problem when handling time-series data for regression tasks.

In general, LSTM-based time-seires prediction models utilize the last hidden state of the sequence to represent the entire input data. This state is then connected to a fully connected layer to generate either a single output (next time step) or multiple outputs (future steps). However, in this research, we concatenate the entire sequence of hidden states and connect them to the multi-output layer. This approach captures more comprehensive past time-series information, enabling effective temporal multi-output regression for real-time wave height prediction by leveraging a broader range of temporal dependencies for improved prediction accuracy.

\subsection{Deep ensemble (DE) for UQ}
\label{subsec:de}
In this study, the DE approach, well-known for its effectiveness in UQ, is applied to overcome the limitations inherent in BNNs, such as robustness and model complexity. DE is a probabilistic model that leverages an ensemble of neural networks (NNs), each generating outputs as a Gaussian distribution $N(\mu(x),\sigma^2(x))$, where $\mu(x)$ and $\sigma^2(x)$ represent the predicted mean and variance, respectively. This framework provides a straightforward and scalable solution for estimating predictive uncertainty in NNs, demonstrating effectiveness in both regression and classification tasks \cite{lakshminarayanan2017simple}. $\sigma(x)$ indicates a measure of model uncertainty, and these probabilistic outputs are calculated using the negative log-likelihood (NLL) loss function, a standard metric for assessing probabilistic model performance.

\begin{equation}
    NLL(\mu(x),\sigma^2(x),y) = -\log(p_{\theta}(y|x)) = \frac{\log\sigma^2(x)}{2} + \frac{(y - \mu(x))^2}{2\sigma^2(x)} + constant
\end{equation}

Despite the strengths of single probabilistic NNs, they are often limited in their ability to estimate only aleatoric uncertainty (data uncertainty), but not epistemic uncertainty (model uncertainty). Aleatoric uncertainty arises from variability inherent to the data, while epistemic uncertainty stems from limited data coverage or lack of knowledge. Since we use real operational data, unlike simulations where only model uncertainty exists due to the absence of data noise, a methodology capable of addressing both data and model uncertainty is essential, as both uncertainties coexist in this case. DE addresses this issue by combining predictions from multiple models, each trained with different random initializations but identical architectures, to capture both types of uncertainty \cite{lakshminarayanan2017simple}. The final predictions are then obtained by aggregating outputs from all NNs, resulting in a Gaussian mixture as follows:

\begin{equation}
\begin{aligned}
    \hat{\mu}(x) &= \frac{1}{M} \sum_{i=1}^{M} \mu_{i}(x) \\
    \hat{\sigma}^2(x) &= \frac{1}{M} \sum_{i=1}^{M} \left( \sigma_{i}^2(x) + \mu_{i}^2(x) \right) - \hat{\mu}^2(x)
\end{aligned}
\label{eq:gaussian_mixture}
\end{equation}
where $M$ denotes the number of probabilistic NNs used in the DE. The aggregated predictive mean of DE is $\hat{\mu}$, and the aggregated predictive uncertainty is $\hat{\sigma}^2$.

\begin{figure*}[htb!]
    \centering
        \includegraphics[width=.95\textwidth]{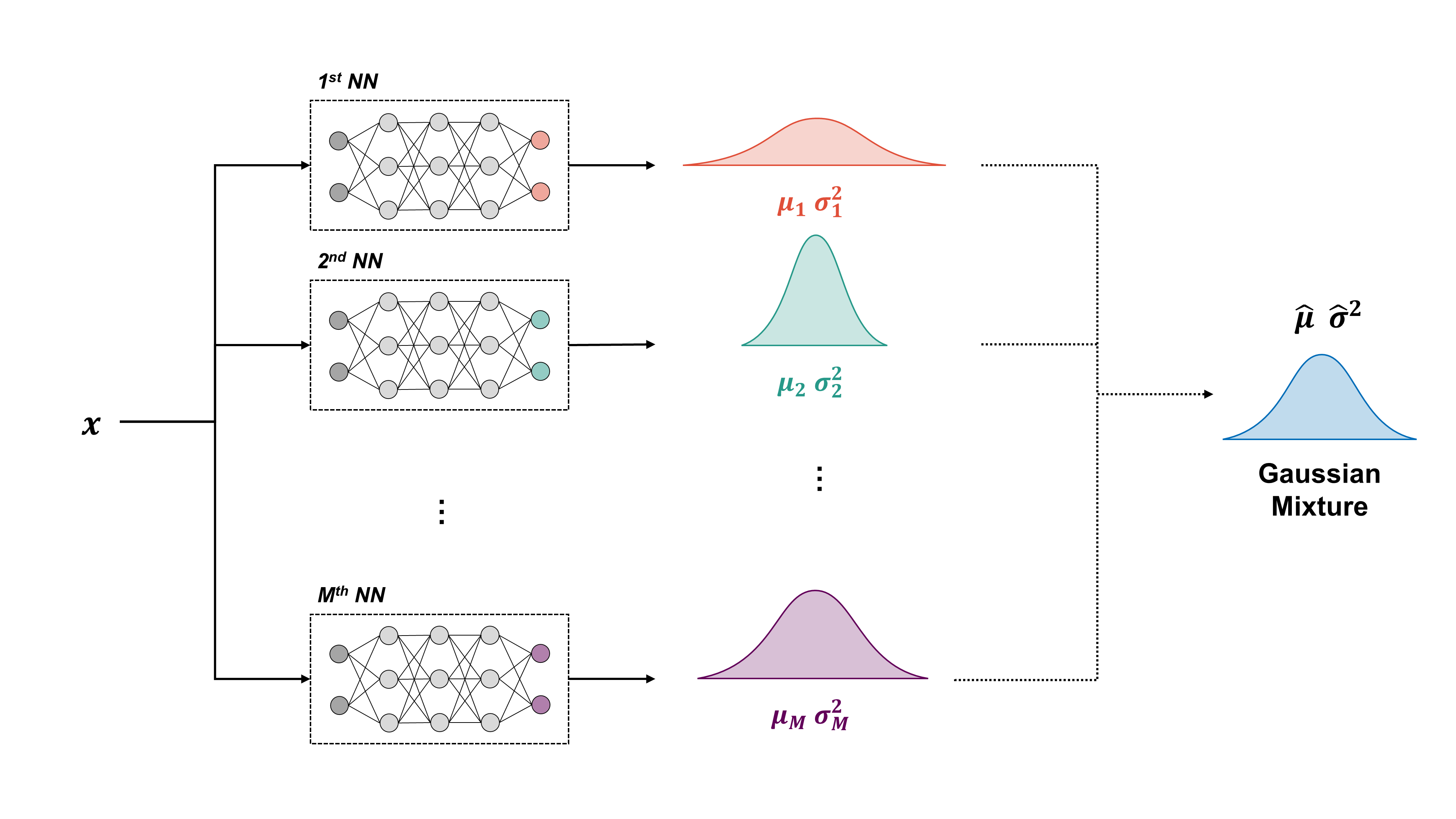}        
    \caption{Architecture of DE approach.}
    \label{fig:deep_ensemble_architecture}
\end{figure*}

The overall DE architecture is illustrated in Fig. \ref{fig:deep_ensemble_architecture}. The primary advantage of DE is the ability to enhance both predictive accuracy and uncertainty estimation without the complex training procedures required by BNNs \cite{fort2019deep}. BNN, which learn distributions over the hyperparameters of the NNs, are theoretically well-motivated by Bayesian principles, but do not perform as well as DEs in practice, particularly under dataset shift. In contrast to BNNs, which are constrained by complex model architectures, DE enables parallel computations without such restrictions. By leveraging model diversity through random initialization of each ensemble component, DEs explore diverse data distributions, enhancing model robustness. Previous studies demonstrate that DE consistently outperforms methods such as Monte Carlo dropout and BNNs in producing well-calibrated uncertainty estimates \cite{lakshminarayanan2017simple, fort2019deep}. Furthermore, DEs are easily scalable to large datasets and complex architectures related to practical applications involving UQ.

Based on these advantages, in this research, the DE method is utilized for temporal multi-output regression in reliable real-time wave height prediction. Leveraging real-world sensor data from operational WEC systems, which inherently includes both aleatoric uncertainty (data uncertainty) and epistemic uncertainty (model uncertainty) due to the irregular and dynamic nature of ocean environments, this approach effectively addresses the coexistence of uncertainties. The simple and scalable LSTM-DE model provides robust and high-quality uncertainty estimates, facilitating real-time decision-making for stable and efficient WEC operations.

\subsection{Evaluation metrics}
\label{subsec:evaluation_metrics}
In this study, we evaluate the performance of LSTM-DE model using four widely utilized evaluation metrics: root mean squared error (RMSE), mean absolute percentage error (MAPE), coefficient of determination (\(R^2\)), and area under the calibration error (AUCE). These metrics are selected for their relevance and effectiveness in quantifying both the predictive accuracy and the quality of UQ in regression tasks.

\subsubsection{Prediction performance}
The prediction performance metrics (RMSE, MAPE, and \(R^2\)) evaluate the accuracy of the predicted values calculated by the LSTM-DE model against the ground truth. These metrics are essential for assessing how effectively the LSTM-DE model captures irregular wave height patterns in regression tasks.

\paragraph{Root mean squared error (RMSE)}
The RMSE is a standard measure of the average magnitude of the prediction error. It is defined as the square root of the mean of the squared differences between the predicted values (\(\hat{y}_i\)) and the actual values (\(y_i\)):

\begin{equation}
    \text{RMSE} = \sqrt{\frac{1}{n} \sum_{i=1}^{n} (y_i - \hat{y}_i)^2}
\end{equation}
where \(n\) represents the number of data points. RMSE is particularly sensitive to large errors, making it a useful metric when the cost of large errors is high.

\paragraph{Mean absolute percentage error (MAPE)}
The MAPE measures the accuracy of the model as a percentage. It is defined as the average of the absolute percentage errors between the predicted and actual values:

\begin{equation}
    \text{MAPE} = \frac{100\%}{n} \sum_{i=1}^{n} \left| \frac{y_i - \hat{y}_i}{y_i} \right|
\end{equation}
where \(n\) represents the number of data points. MAPE provides an intuitive interpretation of prediction accuracy, where lower values indicate better model performance. However, it is important to note that MAPE can be biased if the actual values (\(y_i\)) are close to zero.

\paragraph{Coefficient of determination (\(R^2\))}
The \(R^2\) quantifies the proportion of the variance in the dependent variable that is predictable from the independent variables. It is calculated as follows:

\begin{equation}
    R^2 = 1 - \frac{\sum_{i=1}^{n} (y_i - \hat{y}_i)^2}{\sum_{i=1}^{n} (y_i - \bar{y})^2}
\end{equation}
where \(\bar{y}\) is the mean of the observed data. \(R^2\) values range from 0 to 1, with higher values indicating better model fit. An \(R^2\) of 1 indicates that the model perfectly explains the variance in the data.

\subsubsection{Quality of UQ}
In contrast, the UQ quality metric (AUCE) is employed to evaluate whether the uncertainty estimates calculated by the LSTM-DE model are both reliable and well-calibrated, ensuring they accurately reflect the actual uncertainty associated with the predictions.

\paragraph{Area under the calibration error (AUCE)}
The AUCE is a widely used metric that indicates confidence in the reliability of estimated uncertainty by probabilistic models through a reliability plot, also known as calibration plot \cite{kendall2017uncertainties, kuleshov2018accurate}. The purpose of AUCE measurement is to ensure that the estimated CIs are practically accurate. The mathematical form is as follows:

\begin{equation}
    \text{AUCE} = \frac{1}{K} \sum_{i=1}^{K} \left|\hat{p}_i - p_i\right|
\end{equation}
where $K$ denotes the number of CI candidates. $p_i$ is each CI candidate in $\mathbf{p} = (p_1, p_2, ... p_{K})$, representing the predicted CI. $\hat{p}_i$ is the ratio of instances where the ground truth falls within each CI $p_i$, reflecting the observed CI.

\begin{figure*}[htb!]
    \centering
        \includegraphics[width=.95\textwidth]{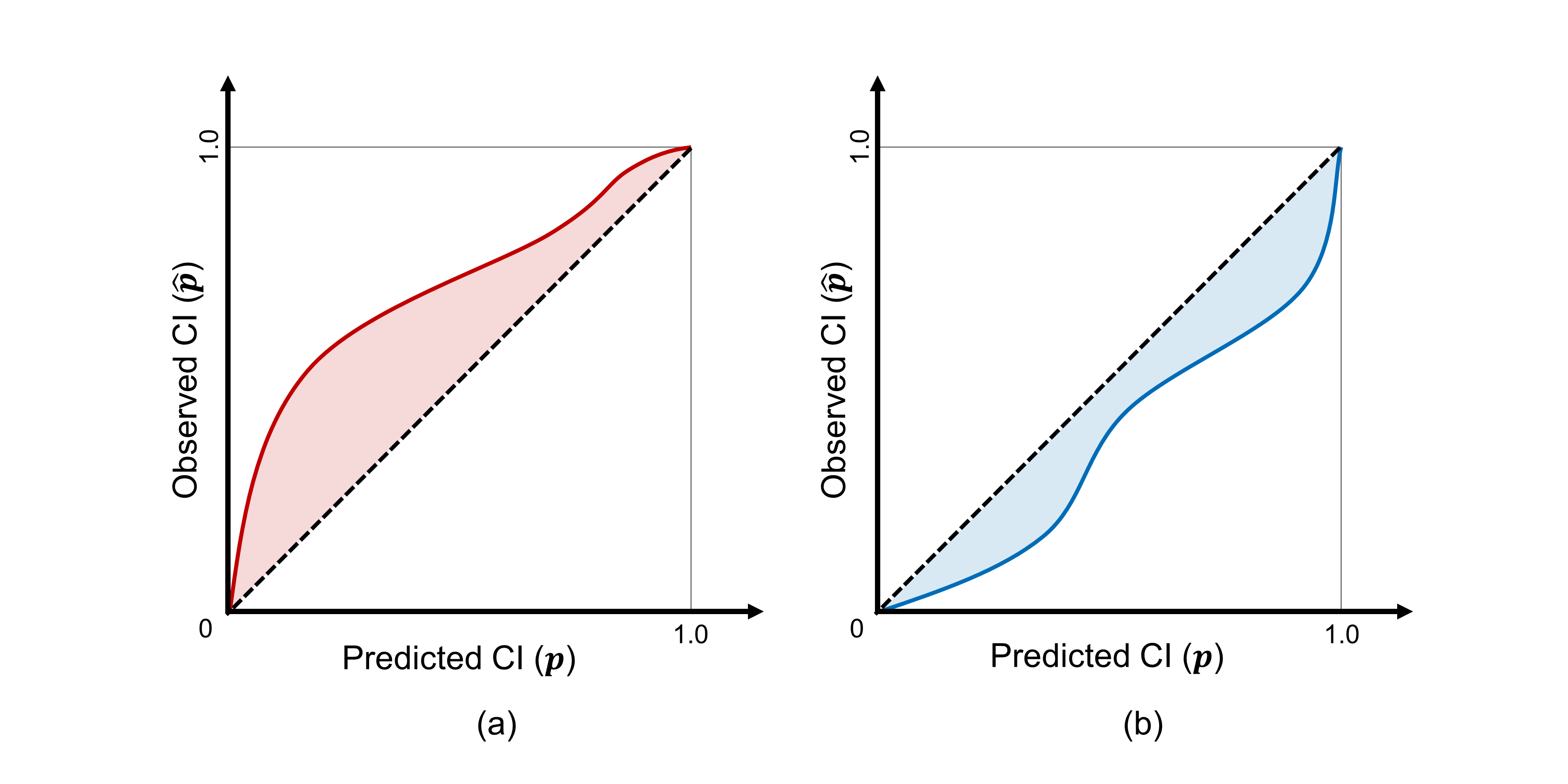}        
    \caption{Example of CI-based reliability plot (a) underconfident and (b) overconfident.}
    \label{fig:reliability_plots}
\end{figure*}

If the observed CI $\hat{p}_i$ exceeds the predicted CI $p_i$, as shown in Fig. \ref{fig:reliability_plots} (a), the model is underconfident. Otherwise, it is overconfident like Fig. \ref{fig:reliability_plots} (b). A well-calibarated model should satisfy $\hat{p}_i \cong p_i$ and have a low AUCE value, indicating that the estimated uncertainty is reliable. Based on the reliability plots for the probabilistic model, AUCE is measured as the area between the calibration curve and ideal line ($\hat{p}_i = p_i$), quantifying the quality of model uncertainty. The tendency of the models to be underconfident or overconfident is assessed by the location of calibration curve and the size of these areas.

\subsection{Standard deviation (STD) scaling for uncertainty calibration}
\label{subsec:std_scaling}
When the estimated uncertainty is imprecise as indicated by AUCE, calibration methods become essential to enhance the quality of UQ. One widely used method is temperature scaling, a parametric approach easily adaptable to engineering application \cite{guo2017calibration}. For regression tasks, a method known as STD scaling was introduced, offering a straightforward yet effective approach to improving uncertainty calibration by adjusting the predicted STD \cite{levi2022evaluating}. In STD scaling, a scaling factor $s$ is optimized and applied to the STD estimated by DE models to minimize the NLL loss, as below:

\begin{equation}
    s = \argmin_{s} \left( \frac{\log(s\hat{\sigma}(x))^2}{2} + \frac{(y - \hat{\mu}(x))^2}{2(s\hat{\sigma}(x))^2} + constant \right)
\end{equation}
where $s$ is the scaling factor. $\hat{\mu}(x)$ and $\hat{\sigma}(x)$ represent the predicted value and estimated uncertainty from the model before calibration, respectively.

This method is a post-process calibration, adjusting only the estimated uncertainty independent from the model training procedure. For this calibration, a separate validation dataset, distinct from train and test datasets, is required. In summary, this study utilizes a straightforward STD scaling method for uncertainty calibration through the scaling factor $s$, while maintaining the integrity of pre-trained NNs. By leveraging STD scaling, the method provides reliable and well-calibrated uncertainty estimates, accurately representing the uncertainty associated with the real-time wave height predictions.

\section{Proposed method: LSTM-DE}
\label{sec:lstm_de_architecture}

\subsection{Necessity of the proposed method}
Reliable real-time wave height prediction requires not only high predictive accuracy but also reliable quantified uncertainties to enable effective decision-making for stable WEC operations. Conventional time-series prediction methods often focus on deterministic outputs, limiting their ability to address the inherent complexities of highly nonlinear and irregular wave behaviors. Existing LSTM-based approaches typically rely on the last hidden state, which fails to leverage the full temporal dynamics of the input sequence. In general, probabilistic models such as BNNs, while promising for UQ, are prone to overfitting due to their tendency to converge to single-mode solutions and require computationally expensive training, making them impractical for real-time applications. Moreover, previous studies, including BNN-based wave height prediction methods \cite{zhang2022phase}, have not employed calibration techniques to improve the quality of uncertainty estimates for model reliability.

The proposed LSTM-DE model architecture addresses these challenges by integrating LSTM networks for temporal prediction with DE for robust UQ. By utilizing all hidden states across the input sequence, the architecture captures a broader range of temporal dependencies, significantly enhancing prediction performance. This approach is similar to the attention mechanism \cite{chorowski2014end, chorowski2015attention}, which also leverage all hidden states; however, the LSTM-DE model assigns equal weights to each state, maintaining simplicity while maximizing effectiveness for model training. DE further enhances the architecture by aggregating predictions from an ensemble of models, each trained with different random initializations, thereby reducing variance and improving the reliability of probabilistic outputs. Additionally, the inclusion of a post-hoc calibration step using STD scaling further aligns predicted uncertainties with observed values, ensuring well-calibrated and reliable uncertainty estimates crucial for real-time decision-making. By combining these novel approaches, the LSTM-DE model offers a robust, simple, scalable, and efficient solution for real-time wave height prediction, ensuring both high predictive accuracy and reliable UQ.

To the best of our knowledge, this is the first study to integrate LSTM utilizing all hidden states, DE for robust UQ, and STD scaling for post-hoc uncertainty calibration into a unified framework for real-time wave height prediction, effectively addressing the limitations of conventional methods. Key advantages of the proposed method are summarized as follows:
\begin{enumerate}
    \item Enhanced LSTM with entire hidden states: Conventional LSTM models rely only on the last hidden state for predictions. The proposed method concatenates all hidden states across the input sequence to capture more comprehensive past temporal information. This approach enables effective temporal multi-output regression, improving predictive accuracy.
    \item Robust, simple, and scalable UQ with DE: DE enhances model robustness by leveraging diverse predictions from multiple models trained with different random initializations for model parameters. By aggregating these predicted values, the simple and scalable DE achieves improved reliability and reduced variance in probabilistic outputs, making it well-suited for scenarios where uncertainty estimation is critical.
    \item Post-hoc uncertainty calibration with STD scaling: STD scaling adjusts predicted uncertainties through a post-processing step, independent of training. This calibration technique aligns predicted uncertainty with observed values using an optimized scaling factor. It ensures reliable uncertainty estimates, crucial for real-time decision-making in dynamic ocean environments.
\end{enumerate}

\subsection{LSTM-DE for temporal multi-output regression with uncertainty}
\label{subsec:lstm_de_architecture}
\begin{figure*}[htb!]
    \centering
        \includegraphics[width=.95\textwidth]{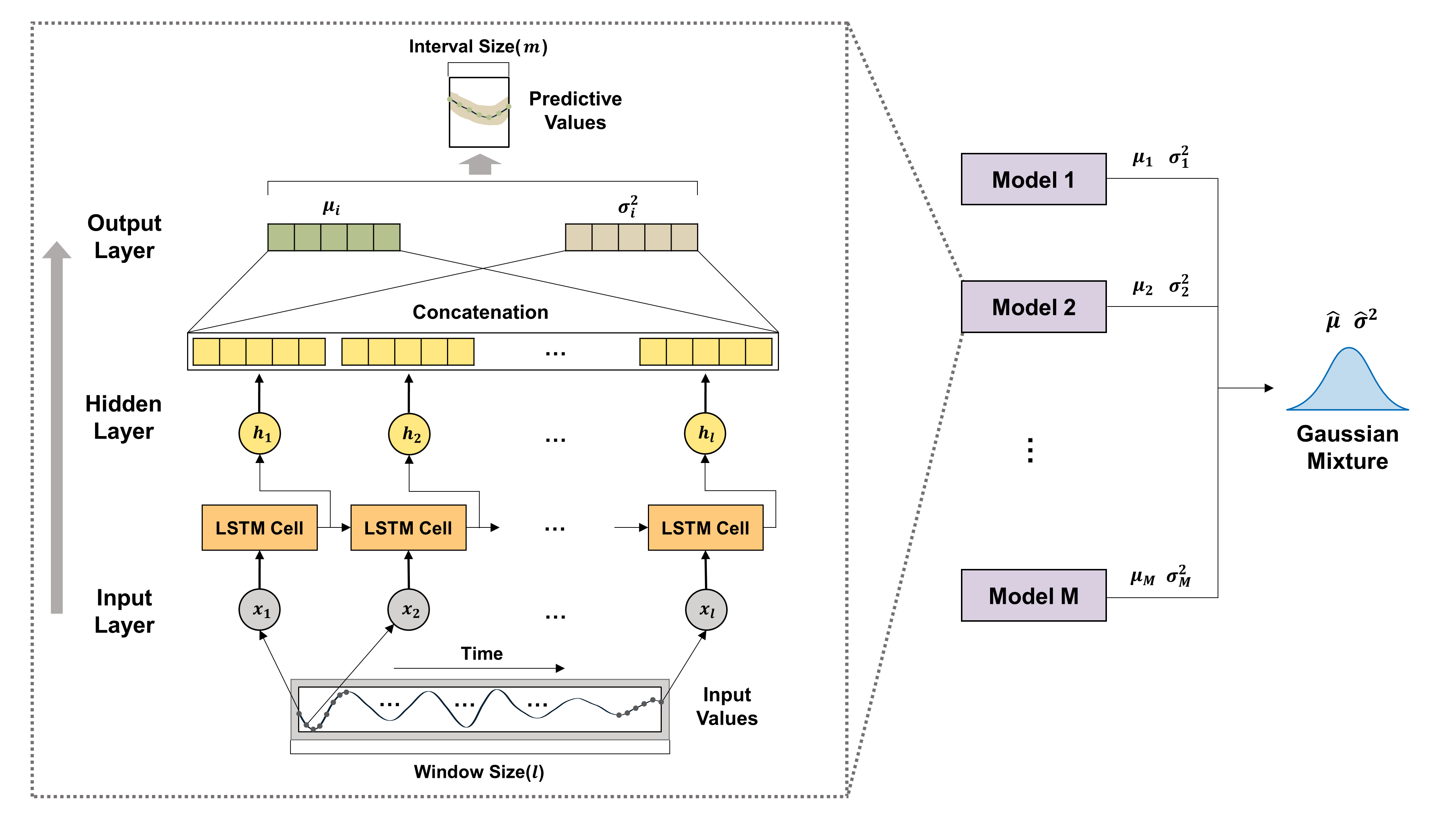}        
    \caption{LSTM-DE model architecture for temporal multi-output regression.}
    \label{fig:lstm_de_architecture}
\end{figure*}

The proposed LSTM-DE model architecture, illustrated in Fig. \ref{fig:lstm_de_architecture}, combines LSTM with the DE approach to develop a lightweight yet powerful probabilistic wave height prediction model. This architecture is designed to balance predictive accuracy and robust UQ. The input data is a time-series sequence $\mathbf{x} = (x_1, x_2, \dots, x_{l})$, fed into the LSTM networks with a window size \(l\), which captures past temporal features across multiple time steps. Each input passes through the LSTM cells, generating the hidden state sequence $\mathbf{h} = (\mathbf{h_1}, \mathbf{h_2}, \dots, \mathbf{h_{l}})$. To maintain the lightweight structure of the model, characterized by a minimized total number of hyperparameters for real-time prediction, a single hidden layer is selected for the LSTM model architecture using concatenation. The output layer consists of two parallel fully connected layers: one predicting the mean ($\mu_{i}$) of the wave height patterns, and the other predicting its variance ($\sigma_{i}^2$). This enables the model to provide not only point estimates but also the uncertainty estimates associated with each prediction, $\sigma_{i}^2$. In general, auto-regressive methods, which are commonly used for time-series predictions and typically aim to predict single future time-step, often encounter the issue of cumulative errors as they sequentially use their own predictions as inputs for subsequent steps \cite{salinas2020deepar, kim2024physics}. This can lead to significant errors over time, particularly in highly nonlinear systems like WECs. To address this issue, a multi-output regression approach was adopted, where the model is trained to predict the consecutive future trajectory $\mathbf{y} = (y_1, y_2, \dots, y_{m})$ following the past window, with a step size equal to the interval size \(m\), rather than focusing on a single time-step regression task. This approach captures a broader scope of temporal information, allowing for more accurate long-term predictions without the drawbacks of auto-regressive methods.

To further enhance the robustness and reliability of the predictions, the DE methodology was integrated into the architecture. DEs offer a straightforward and scalable solution for UQ, as demonstrated in prior studies \cite{lakshminarayanan2017simple, fort2019deep}. This method not only produces uncertainty estimates but also strengthens the model’s robustness, especially when encountering out-of-distribution data. In the proposed architecture, \(M\) independent LSTM models are trained with identical architectures but with different random initializations for model parameters: the use of diverse random initializations across the ensemble allows the model to capture various aspects of the data, thus improving both accuracy and UQ. Finally, each model independently generates predictions $(\mu_{1}, \mu_{2}, \dots, \mu_{M})$ and their uncertainties $(\sigma_{1}^2, \sigma_{2}^2, \dots, \sigma_{M}^2)$, capturing the inherent model uncertainty called epistemic uncertainty. By aggregating predictions from multiple models, the architecture provides a comprehensive estimates of the prediction $(\hat{\mu})$ and uncertainty $(\hat{\sigma}^2)$ in Eq. \eqref{eq:gaussian_mixture}. This enables the model to produce point estimates and their confidence intervals (CIs), ensuring robust prediction and UQ. In general, the ensemble approach is known to significantly enhance the prediction performance and ensures effective UQ, contributing to the overall scalability and robustness of the prediction model. Moreover, as detailed in Section \ref{subsec:std_scaling}, the proposed LSTM-DE model improves the reliability of UQ by incorporating a post-hoc calibration technique called STD scaling. This method aligns the predicted uncertainty with the observed uncertainty by calibrating the CIs through the multiplication of an optimized scaling factor $s$. By enhancing the quality of uncertainty estimates, this additional step strengthens the applicability of the LSTM-DE model architecture for real-time decision-making in dynamic ocean environments, where reliable uncertainty is essential for stable WEC operations.

In summary, the proposed LSTM-DE model architecture effectively combines the temporal modeling strengths of LSTM networks with the robust UQ provided by DE, further enhanced through post-hoc uncertainty calibration using STD scaling. By leveraging LSTM’s ability to capture temporal dependencies across multiple time steps and employing DE to deliver robust and reliable UQ through variance reduction via model diversity, the model achieves enhanced probabilistic time-series predictions. This combination is particularly suited for spatiotemporal regression tasks, such as real-time wave height prediction in dynamic ocean environments, where both predictive accuracy and reliable UQ are essential. To further improve the reliability of UQ, STD scaling is applied as a post-hoc uncertainty calibration method. This technique aligns the predicted uncertainty with observed values by applying an optimized scaling factor $s$, ensuring that the predicted CIs are well-calibrated for real-time decision-making in dynamic and unpredictable ocean environments. By integrating LSTM for enhanced temporal feature extraction, DE for robust, simple, and scalable UQ, and STD scaling for post-hoc uncertainty calibration, the proposed LSTM-DE architecture provides an effective and efficient solution for complex time-series data prediction challenges, particularly in WEC systems.

\section{Results and discussion}
\label{sec:results_and_discussion}
In this section, the performance of the proposed LSTM-DE model is evaluated in terms of probabilistic prediction and reliability. Section \ref{subsec:details_of_experiments} outlines the dataset details and experimental setup, including data preprocessing, train-validation-test splits, and hyperparameter configurations. Subsequently, the baseline model was assessed in Section \ref{subsec:baseline_model} using established scoring metrics. A parametric study, as presented in Section \ref{subsec:parametric_study}, explored the impact of prediction length, window size, interval size, and the number of ensemble models on predictive accuracy and the quality of UQ. Additionally, the model's robustness was validated across diverse ocean wave types, including regular, amplifying, damping, and calm waves, as detailed in Section \ref{subsec:validation_wave_type}, demonstrating its ability to adapt to various wave dynamics while maintaining high predictive accuracy and reliable UQ. Finally, Section \ref{subsec:std_scaling_result} confirmed that STD scaling effectively enhances the quality of UQ, ensuring the model is well-calibrated.

\subsection{Details of experiments}
\label{subsec:details_of_experiments}
\begin{figure*}[htb!]
    \centering
        \includegraphics[width=.95\textwidth]{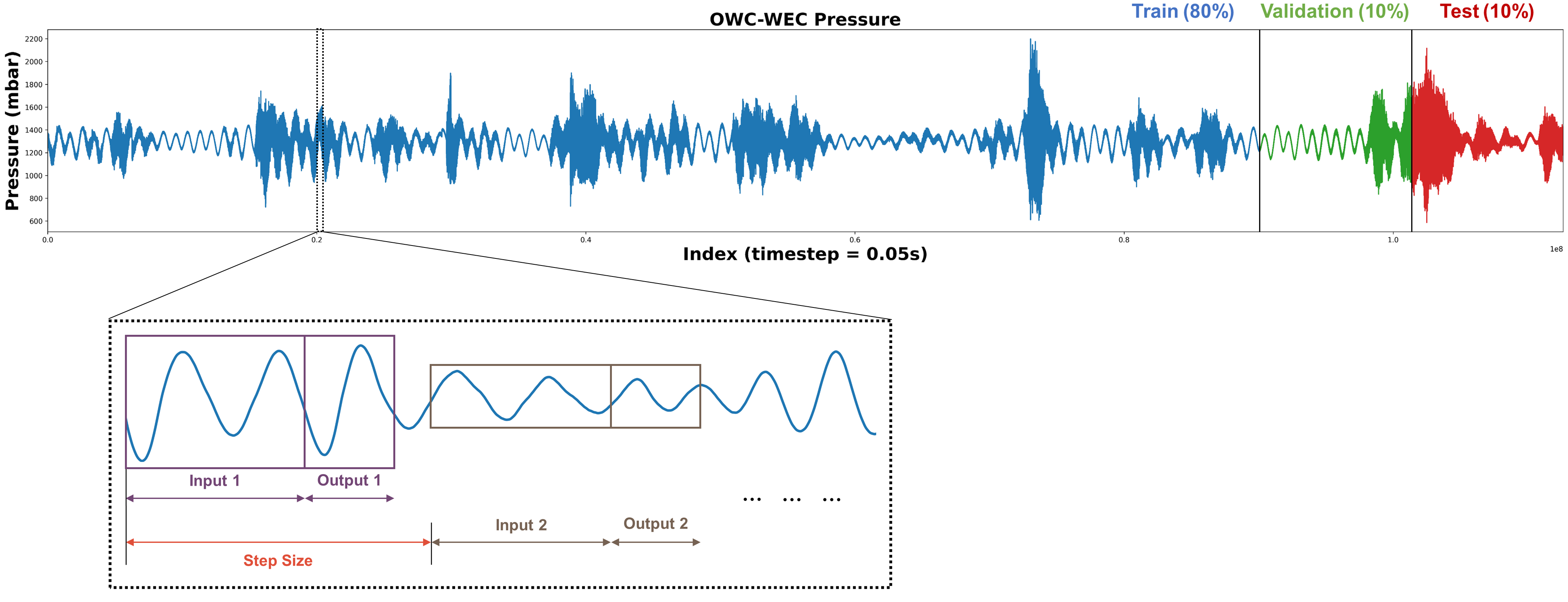}        
    \caption{OWC-WEC operational data and its pre-processing.}
    \label{fig:operational_data_and_slicing_dataset}
\end{figure*}

In this study, the real operational pressure data measured by the 500kw OWC-WEC shown which is installed on the coast of Jeju Island, South Korea in Fig. \ref{fig:owc_wec_jeju}, was utilized to analyze the irregular ocean wave height variations within the chamber. The data was collected over a specific period from December 2020 to February 2021, capturing pressure fluctuations at the bottom of the OWC-WEC chamber with a sampling frequency of 20Hz (0.05 seconds for 1 timestep). It focused primarily on winter conditions, which are expected to provide suitable operational scenarios for wave energy conversion \cite{hong2021response}. As mentioned in Section \ref{subsec:owc_wec}, these pressure values, initially recorded in millibars (mbar), were converted to meters (m) to predict wave heights on a meter scale, based on the principles of hydrostatic pressures. Additionally, min-max normalization was applied to appropriately scale the data. This dataset was split into train (80\%), validation (10\%), and test (10\%) based on time sequence, as illustrated in Fig. \ref{fig:operational_data_and_slicing_dataset}. For UQ calibration, a post-processing step with STD scaling, a separate validation dataset independent of the training and test process, is required \cite{levi2022evaluating}. Therefore, the validation dataset is specifically utilized for this purpose.

As depicted in Fig. \ref{fig:operational_data_and_slicing_dataset}, the dataset was processed to generate input and label datasets for model training through slicing techniques. Specifically, the window size (input) defines the range of past data utilized by the model to understand irregular wave height patterns, while the interval size (output) indicates the range of future data that the model predicts. The step size determines the sampling period of data extraction. By adjusting these hyperparameters (window, interval, and step size), we have explored the model performance, balancing predictive accuracy with computational efficiency. The collected time-series data was preprocessed to generate train, validation, and test datasets, with 80\%, 10\%, and 10\% split, respectively. To ensure robust and reliable performance across different wave scenarios for conducting real-time wave height prediction, min-max normalization was applied to scale the time-series data before training the model. Additionally, the predictive accuracy and the quality of UQ were first assessed using the validation dataset to verify alignment with real-world conditions. The model's final performance was then evaluated on the test dataset. All experiments were conducted on a NVIDIA RTX A6000 GPUs.

\subsection{Baseline model with ocean-related domain knowledge}
\label{subsec:baseline_model}
In this study, a baseline model was constructed and evaluated, with a focus on examining the influence of ocean-related theory and empirical domain knowledge in designing efficient initial experiments to minimize training costs. Following this, a parametric study was conducted to examine the effects of window size, interval size, and the number of models on the predictive accuracy and the quality of UQ. The initial experimental conditions, detailed in Section \ref{subsec:details_of_experiments}, including the key model hyperparameters, are summarized in Table \ref{tab:intial_experimental_conditions}. The baseline LSTM-DE model architecture comprises a single LSTM layer with 70 nodes, trained over 3,500 epochs using the Adam optimizer and early stopping to prevent overfitting. The NLL loss function was employed to ensure accurate quantification of both predictive accuracy and uncertainty. Key hyperparameters, such as step size (50 timesteps: 2.5 seconds), window size (300 timesteps: 15 seconds), and interval size (70 timesteps: 3.5 seconds), were carefully determined through preliminary experiments and informed by ocean-related domain knowledge \cite{kinsman1984wind, falnes2020ocean, previsic2021ocean}.

Ocean waves, driven by wind, can be classified into various categories based on parameters such as frequency and period, ranging from tidal waves (12–24 hours) to capillary waves (less than one second) \cite{kinsman1984wind}. For OWC-WECs, gravity waves, with periods between 1 and 30 seconds, are the most significant, with energy typically concentrated in the 5–15 second range \cite{falnes2020ocean}. To achieve accurate predictions, utilizing past data spanning 2–3 dominant wave periods is crucial \cite{previsic2021ocean}. Furthermore, in winter, which is expected to provide suitable operational scenarios for wave energy conversion conditions in South Korea, waves with periods between 3 and 7 seconds are common. To prevent system shutdowns and avoid potential damage to the system, predicting future wave peaks in real-time is crucial. Therefore, it requires capturing the half-period of waves common in winter: that is, from 1.5 to 3.5 seconds. Based on these research findings, it is feasible to design the model architecture according to the period of gravity wave, guiding the selection of the appropriate length of past and future time-series data.

Using these conditions, the baseline model was constructed and trained on datasets divided into train, validation, and testing sets to ensure robust and reliable performance. Training was repeated across datasets to analyze the model’s performance consistently, leveraging domain knowledge to ensure stability. The baseline settings were established to evaluate predictive accuracy, uncertainty estimates, and the effectiveness of STD scaling for UQ calibration. Further experiments investigated the impact of window size, interval size, and the number of ensemble models compared to the baseline. Additionally, the model's ability to generalize across various wave scenarios was tested and visualized, confirming its robustness and practical applicability for real-time OWC-WEC operations.

\begin{table}[h]
    \centering
    \caption{Initial experimental conditions.}
    \label{tab:intial_experimental_conditions}
    \begin{tabular}{ccccc}
        \toprule
        \textbf{Type} & \textbf{Models} & \textbf{Layer / Nodes} & \textbf{Epochs} & \textbf{Loss Function} \\ \midrule
        LSTM-DE & 5 & 1 / 70 & 3500 & NLL \\ \midrule
        \textbf{Optimizer} & \textbf{Normalization} & \textbf{Step Size} & \textbf{Window Size} & \textbf{Interval Size} \\ \midrule
        Adam & Min-Max & 50 (2.5s) & 300 (15.0s) & 70 (3.5s) \\ \bottomrule
    \end{tabular}
\end{table}

Table \ref{tab:experimental_results_baseline_total} presents the prediction performance of the baseline model across the train, validation, and test datasets. The total training time was approximately 12,673 seconds (approximately 3.5 hours). The model demonstrates high prediction performance, with all datasets showing \(R^2\) values above 0.9, indicating that the model can explain over 90\% of the variance in the data, reflecting its robustness. As expected, the prediction performance on the test dataset is slightly lower compared to the train and validation datasets, due to the test dataset containing unseen data that the model was not exposed to during training process. The error metrics, RMSE and MAPE, also follow a similar trend, with the lower errors on the train and validation datasets compared to the test dataset. However, the model consistently maintains high performance across all datasets, demonstrating its effectiveness and robustness in generalizing to unseen data. This consistency suggests that the model is well-suited for time-extrapolation scenarios, reinforcing its potential for reliable real-time wave height predictions with acceptable accuracy.

\begin{table}[h]
    \centering
    \caption{Experimental results of baseline model (total).}
    \label{tab:experimental_results_baseline_total}
    \begin{tabular}{ccccc}
        \toprule
        \textbf{Dataset} & \textbf{RMSE} & \textbf{MAPE} & \textbf{R\textsuperscript{2}} & \textbf{Training Time [s]}\\ \midrule
        Train (80\%) & 12.23055 & 0.00466 & 0.97161 & 12673 \\
        Validation (10\%) & 11.11542 & 0.00454 & 0.97990 & - \\
        Test (10\%) & 20.64611 & 0.00913 & 0.91242 & - \\ \bottomrule
    \end{tabular}
\end{table}

Additionally, Table \ref{tab:experimental_results_baseline_each_index} presents the predictive accuracy and quality of UQ for each time step (index) from the multi-output regression results on the test dataset. Each time step corresponds to a specific prediction length, representing a future point of interest defined by the user for the given application. It shows a high predictive accuracy at earlier time steps, with \(R^2\) values exceeding 0.95 for predictions made at 0.5s and 1.0s (indices 10 and 20 time steps). However, as the prediction length extends, a noticeable decrease in \(R^2\) is observed. The error metrics RMSE and MAPE follow a similar trend, with lower errors at shorter time intervals and increasing errors as the time step progresses. The AUCE values, which reflect the quality of UQ, demonstrate no specific trend as the prediction length increases, maintaining consistently robust performance. This indicates that, unlike predictive accuracy, the UQ quality remains stable across varying prediction lengths, highlighting the robustness and reliability of the DE approach in managing uncertainty effectively for time-series prediction.

\begin{table}[h]
    \centering
    \caption{Experimental results of baseline model (each index).}
    \label{tab:experimental_results_baseline_each_index}
    \begin{tabular}{ccccc}
        \toprule
        \textbf{Index} & \textbf{RMSE} & \textbf{MAPE} & \textbf{R\textsuperscript{2}} & \textbf{AUCE} \\ \midrule
        total & 20.64611 & 0.00913 & 0.91242 & 0.05123 \\
        10 (0.5s) & \textbf{6.24294} & \textbf{0.00267} & \textbf{0.99335} & 0.07390 \\
        20 (1.0s) & 14.89808 & 0.00647 & 0.96213 & 0.03667 \\
        30 (1.5s) & 22.79748 & 0.01006 & 0.91127 & \textbf{0.02903} \\
        40 (2.0s) & 27.53285 & 0.01222 & 0.87067 & 0.03682 \\
        50 (2.5s) & 28.83263 & 0.01283 & 0.85827 & 0.03833 \\
        60 (3.0s) & 28.01967 & 0.01249 & 0.86607 & 0.03170 \\
        70 (3.5s) & 28.35321 & 0.01256 & 0.86283 & 0.03201 \\ \bottomrule
    \end{tabular}
\end{table}

\begin{figure*}[htb!]
    \centering
        \includegraphics[width=.95\textwidth]{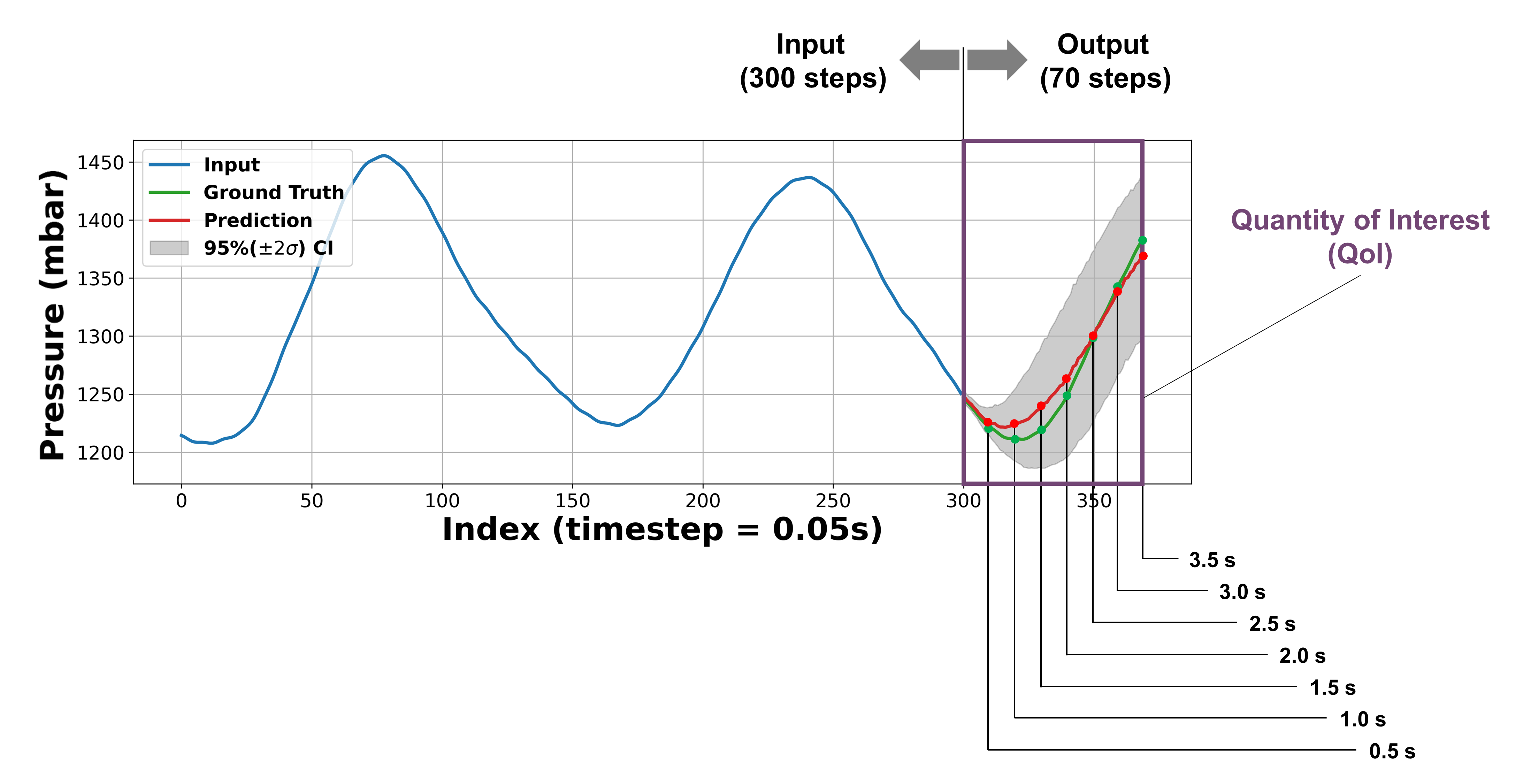}
    \caption{Visualization of temporal multi-output regression using baseline model.}
    \label{fig:experimental_results_baseline_each_index}
\end{figure*}

Fig. \ref{fig:experimental_results_baseline_each_index} illustrates the multi-output prediction of the proposed LSTM-DE baseline model for wave height over time. The prediction is performed over a quantity of interest (QoI) of 70 time steps (3.5 seconds) based on an input sequence length of 300 time steps (15 seconds). Notably, in this study, sequences beyond the QoI (interval size) are deemed out of interest, focusing solely on the interval size specified by ocean-related domain knowledge for real-time wave height prediction. The blue line represents the input data, capturing the past historical wave patterns over 300 time steps. The green line indicates the ground truth for the next 70 time steps, while the red line shows the predicted values by the LSTM-DE. The shaded gray region reflects the 95\% CI for the predicted output, quantifying the uncertainty over the output. In the early time steps, the model demonstrates high accuracy and low uncertainty, particularly in short-term predictions. One of the key observations is the gradual increase in both prediction error and uncertainty as the model forecasts further into the future. In summary, the baseline model performs well in terms of predictive accuracy and UQ at earlier time steps. However, as is typical in time-series forecasting, both prediction errors and CIs increase as the prediction length extends. This highlights the importance of balancing model performance and prediction length when applying this approach to real-time wave energy conversion scenarios, including its integration with facilities and digital twins.

\subsection{Parametric study of baseline model}
\label{subsec:parametric_study}
\begin{figure*}[htb!]
    \centering
        \includegraphics[width=.95\textwidth]{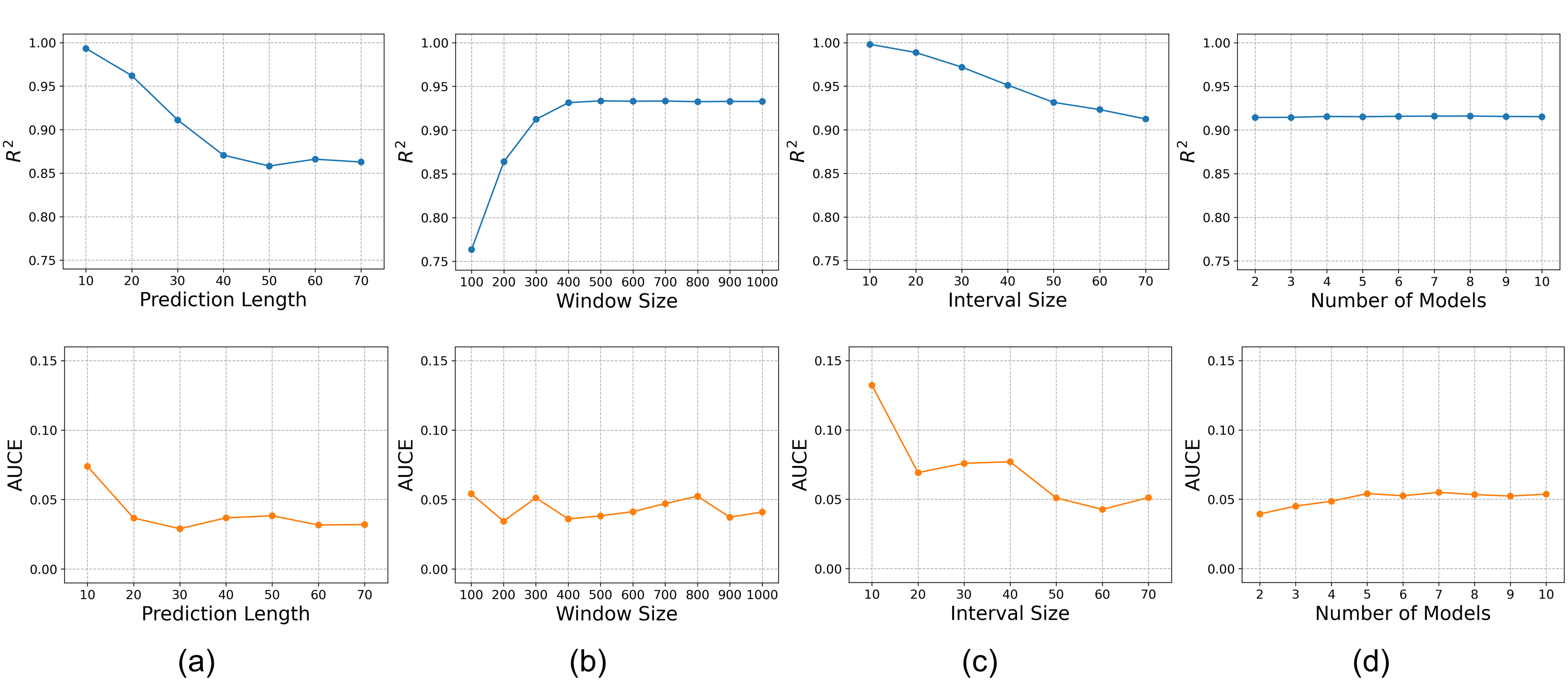}
    \caption{Results of performance evaluation: (a) prediction length, (b) window size, (c) interval size, and (d) number of
models.}
    \label{fig:performance_evaluation_r2_auce}
\end{figure*}

The experimental results, shown in Fig. \ref{fig:performance_evaluation_r2_auce} and Table \ref{tab:experimental_results_various_hyperparameters}, demonstrate the effectiveness of the proposed LSTM-DE model architecture across key model hyperparameters for time-series prediction: prediction length, window size, interval size, and number of models. \(R^2\) and AUCE are used to assess predictive accuracy and the quality of UQ, respectively. This parametric study is expected to provide valuable insights into the trade-offs involved in tuning these hyperparameters for both prediction performance and UQ.

\paragraph{Effects of prediction length}
The effect of the prediction length, which refers to the future wave height prediction point of interest specified by the user for each WEC system, is shown in Fig. \ref{fig:performance_evaluation_r2_auce} (a). In this study, while the LSTM-DE model generates multi-output predictions for the entire interval size of 70 time steps, the prediction length is primarily focused on 50 time steps (2.5 seconds). These values are selectively extracted from the model's output and applied for real-time wave height prediction. As the prediction length increases, the \(R^2\) value steadily decreases but stabilizes after around 40 to 50 time steps. Similarly, the AUCE decreases sharply until about 20 time steps and then shows consistent values with minimal variation. These results indicate that while both predictive accuracy and the quality of UQ decrease over time, they stabilize beyond a certain point.

In summary, the LSTM-DE model demonstrates strong performance for shorter-term predictions with high predictive accuracy. While \(R^2\) decreases as the prediction length increases, the quality of UQ remains stable, with only a modest impact on overall performance. This highlights a trade-off in applications: users can choose to focus on highly accurate short-term predictions or prioritize reasonable accuracy for longer-term predictions, depending on the specific operational requirements of the WEC system.

\paragraph{Effects of window size}
The effect of the window size, which refers to the range of past input data utilized by the model to capture the past wave height patterns that are highly irregular, is shown in Fig. \ref{fig:performance_evaluation_r2_auce} (b). As the window size increases, the \(R^2\) value increases significantly, with a noticeable improvement between 100 and 400 time steps. Beyond 400 time steps, the \(R^2\) stabilizes around 0.95, indicating that increasing the window size up to a certain point allows the model to better capture irregular wave height patterns, thereby improving predictive accuracy. Furthermore, window sizes of approximately 300–400 time steps (15–20 seconds) are suggested as optimal for extracting relevant temporal information in wave height prediction tasks. This choice aligns with the period of gravity waves (1-30 seconds), as detailed in Section \ref{subsec:baseline_model}, where the majority of wave energy is concentrated in the 5–15 second range, equivalent to 2–3 dominant wave periods. By utilizing past data spanning 15–20 seconds, the model effectively captures the most critical wave dynamics, ensuring accurate predictions while minimizing the computational overhead associated with excessively long window sizes. In contrast, the AUCE remains relatively stable across different window sizes, fluctuating slightly around 0.05. This indicates that increasing the window size significantly enhances predictive accuracy without affecting the quality of UQ, which remains stable across various window sizes. However, it is important to note that larger window sizes also increase training time due to the more computational effort required to process longer sequences of time-series data. This trade-off should be considered when selecting the window size, especially for real-time applications where computational efficiency is critical.

In summary, while increasing the window size intuitionally improves predictive accuracy, it comes at the cost of significantly higher memory usage and increased training time, as shown in Table \ref{tab:experimental_results_various_hyperparameters}. Domain knowledge suggests that using approximately 300 time steps (15 seconds) as input is sufficient to achieve strong performance ($R^2 > 0.9$), as it effectively captures the critical wave dynamics within the dominant energy range of gravity waves. Beyond this, larger window sizes may incorporate excessive and less relevant information, leading to diminishing returns in performance while unnecessarily increasing computational demands. Thus, a window size of around 300 time steps strikes an optimal balance between predictive accuracy, uncertainty quality, and computational efficiency, and is recommended for practical applications.

\paragraph{Effects of interval size}
The effect of the interval size, which refers to the range of future label data that the model predicts, is shown in Fig. \ref{fig:performance_evaluation_r2_auce} (c). As the interval size increases, the \(R^2\) value, calculated across all predicted future time steps corresponding to the interval size, steadily decreases, starting from nearly 1.0 at an interval size of 10 time steps and dropping to around 0.9 at an interval size of 70 time steps. This trend indicates that as the interval size increases, the predictive accuracy decrease, capturing broader temporal wave height patterns. On the other hand, the AUCE drops sharply from 0.15 to 0.05 as the interval size increases, stabilizing after 20 time steps. This result suggests that while larger interval sizes reduce predictive accuracy, they contribute to stabilizing UQ.

In summary, increasing the interval size negatively impacts predictive accuracy, as reflected by the decline in \(R^2\), but it enhances the quality of UQ, with the AUCE remaining low and stable at larger intervals. This suggests that predicting longer intervals, while introducing a trade-off in predictive accuracy, allows the model to capture more comprehensive output trends, ultimately improving the reliability of UQ. Conversely, shorter intervals provide higher predictive accuracy but may result in lower UQ quality due to the limited information available for the model to infer broader patterns.

\paragraph{Effects of number of models used in ensemble}
The effect of the number of models, which refers to how many ensemble models are utilized for DE model, is shown in Fig. \ref{fig:performance_evaluation_r2_auce} (d). Interestingly, as the number of models increases from 2 to 10, the \(R^2\) remains relatively stable around 0.9. This result suggests that the beyond a certain point, adding more models does not significantly improve predictive accuracy, indicating that the ensemble achieves sufficient robustness with a smaller number of models. However, the AUCE shows a slight upward trend as more models are added, suggesting that while predictive accuracy remains stable, additional models yield minor instability in UQ. This could be attributed to the increased diversity of predictions from the additional models, leading to slightly higher variance in uncertainty estimation (underconfidence).

In summary, the number of models in the ensemble has a minimal effect on predictive accuracy, as evidenced by the consistent \(R^2\) values across different ensemble sizes. However, the quality of UQ, reflected in the AUCE, shows slight instability with smaller ensemble sizes (2–4 models) due to the stronger influence of individual models. A more stable UQ performance is observed with 5 (or more) models, aligning with the recommendations (5 models) from prior studies \cite{lakshminarayanan2017simple}, which suggested using at least 5 models for robust ensemble predictions. These findings indicate that determining the number of NNs in a DE should prioritize UQ quality considerations over predictive accuracy, given the variability in uncertainty estimation.

\begin{table}[h]
    \centering
    \caption{Experimental results for various hyperparameters.}
    \label{tab:experimental_results_various_hyperparameters}
    \begin{tabular}{ccccccc}
        \toprule
        \multirow{2}{*}{\textbf{Hyperparameters}} && \multicolumn{4}{c}{\textbf{Metrics}} & \multirow{2}{*}{\textbf{Training time [s]}} \\
        \cmidrule{3-6}
        && RMSE & MAPE & R\textsuperscript{2} & AUCE & \\
        \midrule
        \multirow{7}{*}{Prediction length}
        & 10 & \textbf{6.24294} & \textbf{0.00267} & \textbf{0.99335} & 0.07390 \\
        & 20 & 14.89808 & 0.00647 & 0.96213 & 0.03667 \\
        & 30 & 22.79748 & 0.01006 & 0.91127 & \textbf{0.02903} \\
        & 40 & 27.53285 & 0.01222 & 0.87067 & 0.03682 & 12673 \\
        & 50 & 28.83263 & 0.01283 & 0.85827 & 0.03833 \\
        & 60 & 28.01967 & 0.01249 & 0.86607 & 0.03170 \\
        & 70 & 28.35321 & 0.01256 & 0.86283 & 0.03201 \\
        \midrule
        \multirow{10}{*}{Window size}
        & 100 & 32.40151 & 0.01443 & 0.76353 & 0.05418 & 6611 \\
        & 200 & 25.27089 & 0.01100 & 0.86389 & \textbf{0.03444} & 12288 \\
        & 300 & 20.64611 & 0.00913 & 0.91242 & 0.05123 & 12673 \\
        & 400 & 18.38633 & 0.00818 & 0.93147 & 0.03606 & 18977 \\
        & 500 & \textbf{18.06858} & 0.00807 & \textbf{0.93350} & 0.03827 & 19679 \\
        & 600 & 18.11106 & 0.00805 & 0.93306 & 0.04128 & 25011 \\
        & 700 & 18.22137 & \textbf{0.00804} & 0.93331 & 0.04712 & 35198 \\
        & 800 & 18.28604 & 0.00808 & 0.93254 & 0.05238 & 36234 \\
        & 900 & 18.35234 & 0.00809 & 0.93287 & 0.03729 & 33722 \\
        & 1000 & 18.23074 & 0.00807 & 0.93279 & 0.04103 & 39600 \\
        \midrule
        \multirow{7}{*}{Interval size}
        & 10 & \textbf{2.87800} & \textbf{0.00112} & \textbf{0.99815} & 0.13229 & 32946 \\
        & 20 & 6.93955 & 0.00278 & 0.98869 & 0.06919 & 27115 \\
        & 30 & 10.88710 & 0.00460 & 0.97196 & 0.07594 & 19455 \\
        & 40 & 14.56053 & 0.00632 & 0.95128 & 0.07707 & 12137 \\
        & 50 & 17.53366 & 0.00761 & 0.93148 & 0.05100 & 13568 \\
        & 60 & 18.94077 & 0.00837 & 0.93223 & \textbf{0.04271} & 14227 \\
        & 70 & 20.64611 & 0.00913 & 0.91242 & 0.05123 & 12673 \\
        \midrule
        \multirow{9}{*}{Number of models}
        & 2 & 20.45047 & 0.00904 & 0.91440 & \textbf{0.03940} & 5961 \\
        & 3 & 20.42105 & 0.00903 & 0.91447 & 0.04508 & 8508 \\
        & 4 & 20.32272 & 0.00900 & 0.91549 & 0.04860 & 11026 \\
        & 5 & 20.32808 & 0.00899 & 0.91523 & 0.05411 & 13826 \\
        & 6 & 20.24431 & 0.00896 & 0.91571 & 0.05253 & 16830 \\
        & 7 & 20.22056 & 0.00896 & 0.91589 & 0.05501 & 19349 \\
        & 8 & \textbf{20.20750} & \textbf{0.00895} & \textbf{0.91596} & 0.05340 & 22462 \\
        & 9 & 20.26466 & 0.00897 & 0.91552 & 0.05233 & 24584 \\
        & 10 & 20.28958 & 0.00897 & 0.91526 & 0.05368 & 27409 \\
        \bottomrule
    \end{tabular}
\end{table}

\subsection{Validation on various ocean wave types}
\label{subsec:validation_wave_type}
\begin{figure*}[htb!]
    \centering
        \includegraphics[width=.95\textwidth]{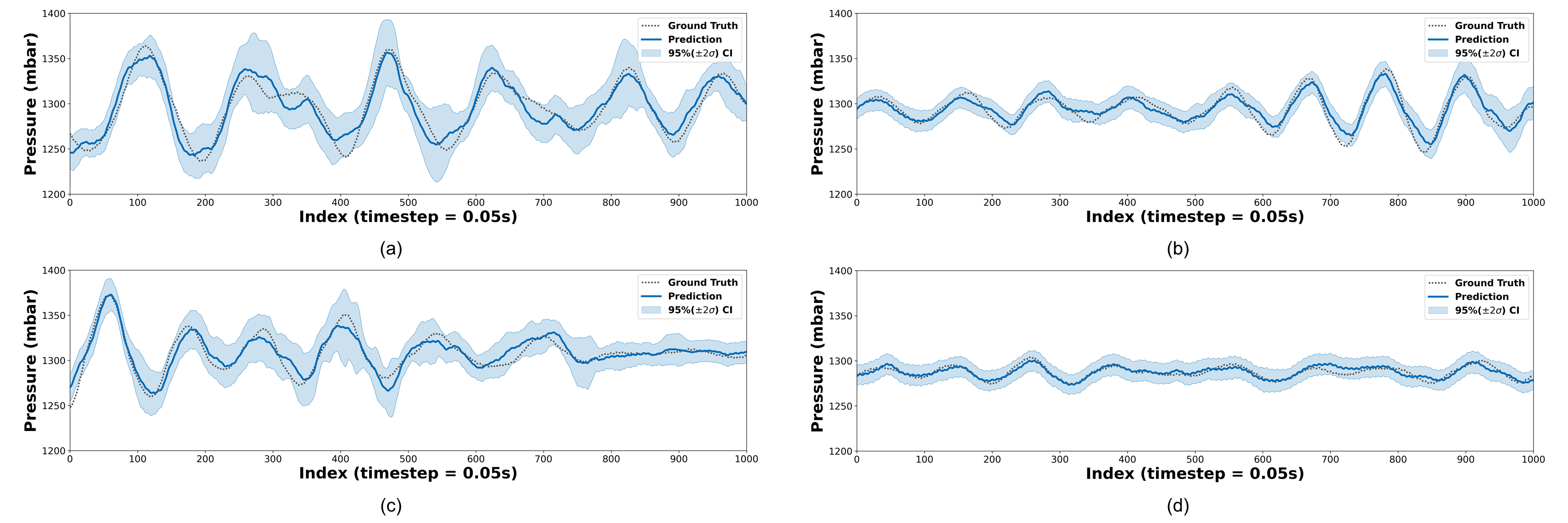}
    \caption{Prediction performance on various ocean wave types (a) regular, (b) amplifying, (c) damping, and (d) calm.}
    \label{fig:various_wave_types}
\end{figure*}

Ocean waves exhibit a variety of motions, characterized by differences in wavelength, amplitude, and period. In this study, we assess whether the proposed LSTM-DE model is able to accurately predict these different types of ocean waves, including regular, amplifying, damping, and calm waves. Fig. \ref{fig:various_wave_types} illustrates the prediction performance on these wave types, highlighting their distinct wavelength, amplitude, and period. These predictions provide insight into the ability of prediction model to capture wave dynamics under different operational conditions while quantifying uncertainty over time. In the context of real-time wave height prediction, the 95\% CI offers uncertainty bounds around the prediction values. The prediction values cover the next 50 steps (2.5 seconds), based on input data spanning 300 steps (15 seconds) of past wave patterns, visualized over a total period of 1,000 steps (50 seconds).

Fig. \ref{fig:various_wave_types} (a) shows regular wave pattern with stable oscillation. The ground truth remains within the 95\% CI, indicating that the LSTM-DE model effectively captures the regularity of ocean waves with high predictive accuracy. Additionally, the results represent that uncertainty increases at the peaks and troughs, where wave height changes more dramatically. This suggests that the predictions are robust during stable wave periods but tend to become more uncertain in areas of rapid change, which can be regarded as reasonable UQ. The two wave patterns shown in Fig. \ref{fig:various_wave_types} (b) and (c) represent amplifying and damping waves over time. For the amplifying wave, the predictive accuracy remains high, with ground truth staying within the 95\% CI. Similarly, for the damping wave, the predictions align well with the ground truth, demonstrating stable performance even as the oscillations diminish. However, the CI widens during wave direction transitions, as shown in Fig. \ref{fig:various_wave_types} (b) and (c), indicating that the uncertainty increases as the wave dynamics change more dramatically. Fig. \ref{fig:various_wave_types} (d) indicates calm wave pattern similar to flat ocean conditions with minimal oscillation. The model achieves higher predictive accuracy compared to the stable, amplifying, and damping waves, with predictions closely aligning with the ground truth. The uncertainty bounds are the narrowest in this case, as the minimal variation in wave height leads to more straightforward predictions and reduced uncertainty. This indicates that the LSTM-DE model performs well when wave height remains relatively constant, maintaining higher predictive accuracy and consistently low uncertainty, as the model is not required to account for rapid changes in wave behavior.

In summary, the results demonstrate that the LSTM-DE model is able to provide accurate predictions and robust UQ across various wave types, including regular, amplifying, damping, and calm waves. The model effectively adapts to different wave dynamics, maintaining high predictive accuracy while appropriately adjusting uncertainty based on the rate of change in wave behavior. This makes it highly applicable to dynamic environments such as WEC systems, where accurate predictions and reliable UQ are essential for ensuring stable and efficient operations. Unlike prior studies that primarily relied on simulation-based or experimental data, this research validates the proposed model using real operational data, demonstrating its robustness and effectiveness in handling the complexities and irregularities inherent in real-world ocean wave. This advancement highlights a significant contribution, setting baselines for practical applicability in WEC systems.

\subsection{STD scaling for UQ calibration}
\label{subsec:std_scaling_result}
\begin{figure*}[htb!]
    \centering
        \includegraphics[width=.95\textwidth]{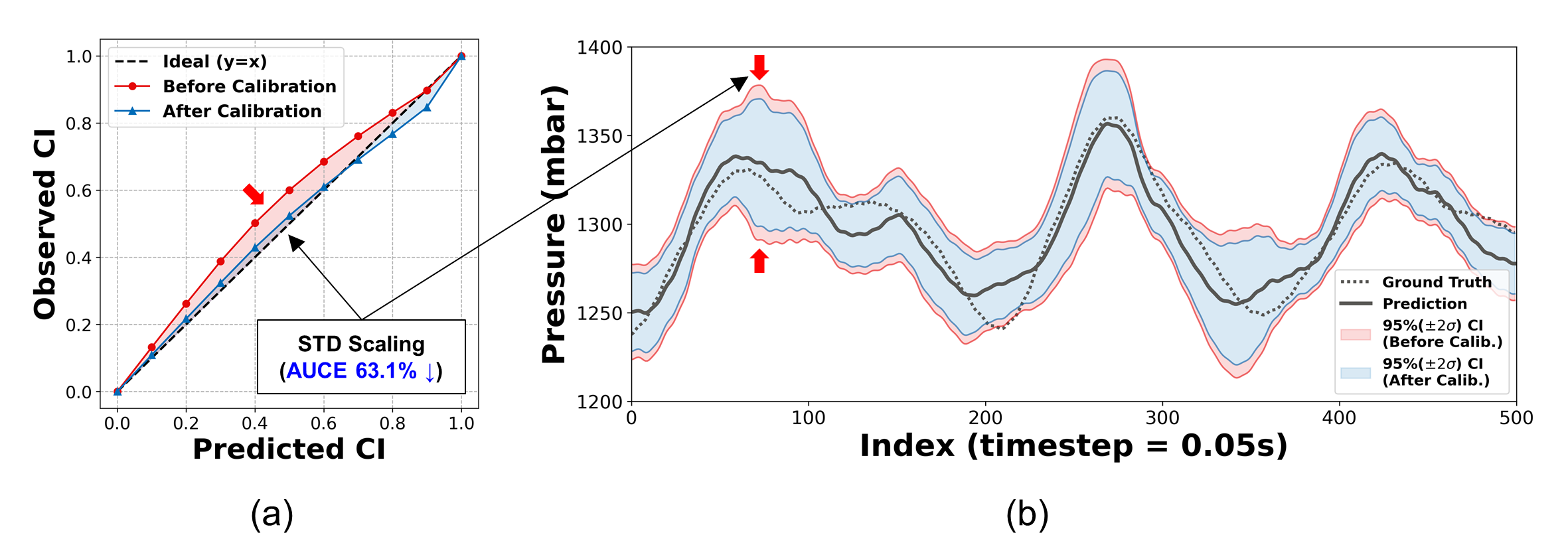}
    \caption{STD scaling of baseline model (a) reliability plot (b) prediction performance.}
    \label{fig:uq_calibration_baseline}
\end{figure*}

In this study, as outlined in Section \ref{subsec:std_scaling}, experiments were conducted to evaluate the effectiveness of STD scaling in improving the quality of UQ for LSTM-DE model. This analysis focuses on the alignment between predicted and observed CIs, using the AUCE as the primary metric to assess calibration performance. Fig. \ref{fig:uq_calibration_baseline} illustrates the impact of STD scaling on the baseline model, comparing performance before and after UQ calibration. In Fig. \ref{fig:uq_calibration_baseline} (a), the red line, representing the model before calibration, shows noticeable deviation from the ideal line (${y} = {x}$) with a tendency for underconfidence. In contrast, the blue line, representing the model after calibration, aligns closely with the ideal line, signifying more reliable UQ with the model CIs better matching observed values---AUCE decreases 63.1\%. Additionally, Fig. \ref{fig:uq_calibration_baseline} (b) shows that STD scaling improves UQ alignment for real-time wave height prediction based on the test dataset. These experimental results demonstrate that STD scaling effectively reduces the AUCE,highlighting an improvement in aligning predicted uncertainties with observed values. This reduction underscores the success of the proposed method in achieving the goal of a reliable prediction model, where uncertainty estimates are well-calibrated for real-world applications.

\begin{figure*}[htb!]
    \centering
        \includegraphics[width=.95\textwidth]{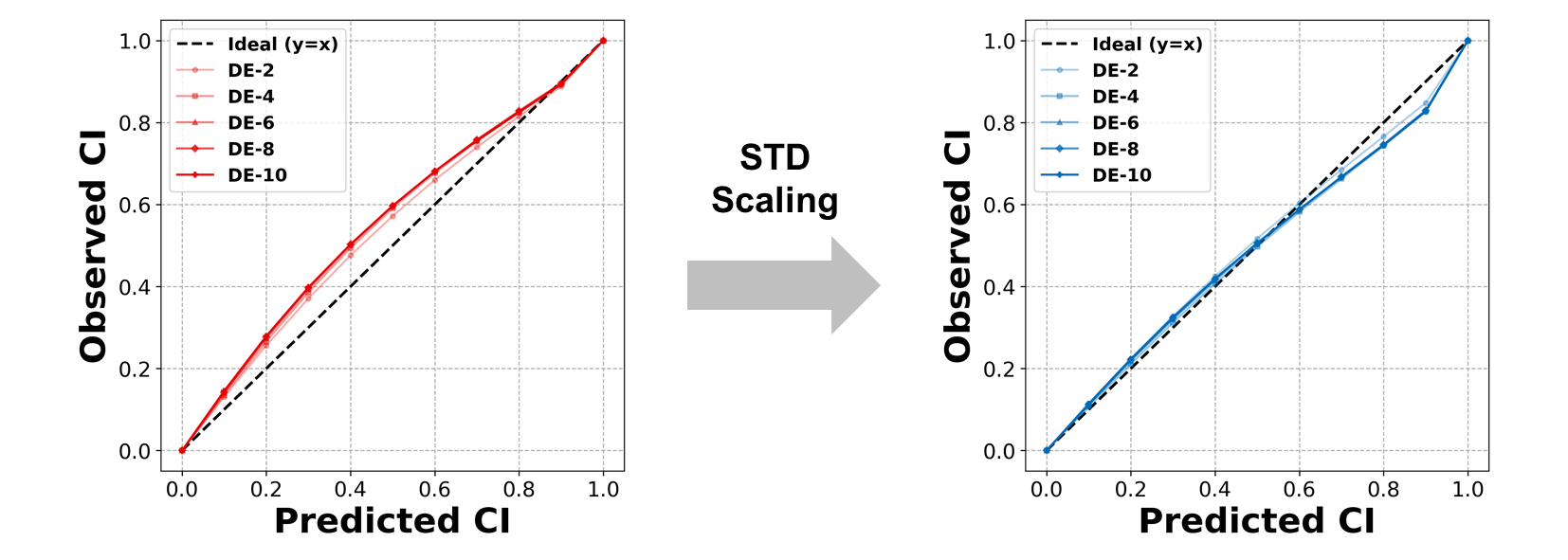}
    \caption{Effects of STD scaling on various numbers of models.}
    \label{fig:uq_calibration_de_models}
\end{figure*}

Quantitatively, the improvement by applying STD calibration on various DE models with different number of NNs are summarized in Table \ref{tab:experimental_results_uq_calibration_de_models}: the decrease on AUCE over 50\% compared to before calibration can be verified. Fig. \ref{fig:uq_calibration_de_models} extends this analysis by examining the effects of STD scaling across different number of models from 2 to 10. In general, ensemble models tend to become underconfident as the number of models increase \cite{rahaman2021uncertainty, yang2024towards}. As shown in Table \ref{tab:experimental_results_uq_calibration_de_models} and Fig. \ref{fig:uq_calibration_de_models}, this expected trend is evident as AUCE clearly increases. Moreover, the results indicates that reliable UQ can still be achieved through STD scaling, regardless of the numbers of models. This capability is essential for ensuring that the CIs in the prediction model remain well-calibrated, thereby enabling dependable probabilistic predictions. Notably, all scaling factors $s$ are less than 1, reflecting the underconfident condition across all DE models (calibration decreases the degree of predictive uncertainty with $s<1$). These well-calibrated reliable prediction models are especially valuable for operational decision-making in uncertain environments like ocean wave predictions, where both accuracy and uncertainty reliability are critical. In conclusion, these findings clearly demonstrate the effectiveness of STD scaling in the LSTM-DE model architecture, with negligible additional computation time required. This study therefore validates the necessity of STD scaling-based uncertainty calibration and strongly recommends its use.

\begin{table}[h]
    \centering
    \caption{Experimental results for STD scaling on various numbers of models.}
    \label{tab:experimental_results_uq_calibration_de_models}
    \begin{tabular}{ccccc}
        \toprule
        \multirow{2}{*}{\textbf{Models}} & \multicolumn{3}{c}{\textbf{AUCE}} & \multirow{2}{*}{\textbf{Scaling factors}} \\ 
        \cmidrule{2-4}
        & Before calibration & After calibration & Variation & \\ 
        \midrule
        DE-2 & \textbf{0.03940} & \textbf{0.01923} & 51.2\% $\downarrow$ & 0.86652 \\
        DE-3 & 0.04508 & 0.02154 & 52.2\% $\downarrow$ & 0.86652 \\
        DE-4 & 0.04860 & 0.02046 & 57.9\% $\downarrow$ & 0.78296 \\
        DE-5 & 0.05411 & 0.02453 & 54.7\% $\downarrow$ & 0.74425 \\
        DE-6 & 0.05253 & 0.02143 & 59.2\% $\downarrow$ & 0.78296 \\
        DE-7 & 0.05501 & 0.02308 & 58.0\% $\downarrow$ & 0.78296 \\
        DE-8 & 0.05340 & 0.02247 & 57.9\% $\downarrow$ & 0.78296 \\
        DE-9 & 0.05233 & 0.02294 & 56.2\% $\downarrow$ & 0.78296 \\
        DE-10 & 0.05368 & 0.02363 & 56.0\% $\downarrow$ & 0.78296 \\
        \bottomrule
    \end{tabular}
\end{table}

\section{Conclusions and future work}
\label{sec:conclusions_and_future_work}
Although WECs represent a promising future power generation system, they face challenges such as low energy density and non-stationary ocean wave conditions. To address these issues, the safe and efficient operation of WEC facilities, supported by a high-performance reliable real-time wave height prediction model, is essential. In this study, we proposed a novel AI-powered reliable real-time wave height prediction model using an LSTM-DE model architecture coupled with STD scaling for post-hoc uncertainty calibration to guarantee high predictive accuracy and model reliability. This approach allows WECs to achieve robust and reliable wave height predictions, even under the highly irregular nature of ocean waves.

Experimental validation using high-fidelity operational data from an OWC-WEC operational data demonstrated high predictive accuracy ($R^2 > 0.9$) over short-term predictions (3.5 seconds) based on past data (15 seconds) while reducing AUCE by over 50\% through STD scaling across different ensemble configurations. These results highlight the effectiveness of the proposed LSTM-DE model in achieving the dual objectives of high predictive accuracy and reliable UQ, underscoring its importance as an effective and efficient solution for reliable wave height prediction. Furthermore, a comprehensive parametric study on model hyperparameters demonstrated the advantages of incorporating ocean-related domain knowledge. For instance, selecting a window size of 300 time steps (15 seconds) aligned with the dominant energy range of gravity waves, achieving an optimal balance between predictive accuracy ($R^2 > 0.9$) and computational efficiency. This study provided valuable insights into the trade-offs involved in hyperparameter tuning, offering practical guidelines for developing reliable prediction models. The proposed LSTM-DE model was also validated across various ocean wave types, including regular, amplifying, damping, and calm waves. The model consistently delivered accurate predictions while effectively quantifying uncertainty, representing its adaptability to diverse wave conditions and its robustness in handling real-world irregularities. By doing so, these findings suggest that the LSTM-DE model architecture can be effectively adapted for various WEC systems, supporting ocean wave prediction across different ocean energy conversion applications. In conclusion, the proposed method shows potential to enhance WEC availability, accelerate digital transformation, and advance the commercialization of ocean energy in digital twin applications.

Future work will focus on three key topics to further enhance the applicability of LSTM-DE model architecture. First, continuous operational data collection and the integration of a feedback loop. Expanding the model to incorporate multi-year datasets, rather than data from a single year, would improve the model learning and applicability. A continuous feedback loop with real-time data updates will allow the AI model to dynamically refine its predictions, improving robustness across varying conditions. Second, model hyperparameter optimization to account for seasonality and operational conditions. Tuning the model hyperparameters for seasonal variations and specific operational conditions is expected to enhance both predictive accuracy and model reliability, as wave patterns shift significantly with the seasons. Finally, minimization of model uncertainty. Further reducing model uncertainty remains an essential goal. Leveraging larger datasets and refining calibration techniques will enhance the model reliability across diverse and extreme wave conditions. Collaboration with domain experts to adjust calibration factors based on operational insights could further improve model reliability, particularly for specific operational conditions and scenarios.

\clearpage

\section*{CRediT authorship contribution statement}
\textbf{D. Lee}: Methodology, Software, Validation, Formal analysis, Investigation, Visualization, Writing – original draft.
\textbf{S. Yang}: Conceptualization, Methodology, Writing – review \& editing.
\textbf{J. Oh}: Data curation, Project administration.
\textbf{S. Cho}: Data curation, Project administration.
\textbf{S. Kim}: Writing – review \& editing.
\textbf{N. Kang}: Supervision, Conceptualization, Writing – review \& editing, Funding acquisition.


\section*{Declaration of competing interest}
The authors declare that they have no known competing financial interests or personal relationships that could have appeared to influence the work reported in this paper.

\section*{Data availability}
The authors do not have permission to share data.

\section*{Acknowledgments}
This research was supported by the Korea Research Institute of Ships and Ocean Engineering, a grant from the Endowment Project “Digital Platform Development to Support Marine Digital Transformation” funded by the Ministry of Oceans and Fisheries (2520000291). Also, this work was supported by grants from the Ministry of Science and ICT of Korea (Nos. 2022-0-00969, 2022-0-00986, GTL24031-300, and RS-2024-00355857) and the Ministry of Trade, Industry \& Energy of Korea (No. RS-2024-00410810).



\bibliographystyle{unsrtnat}
\bibliography{references}  





\end{document}